\PassOptionsToPackage{table}{xcolor}
\documentclass[10pt,twocolumn,letterpaper]{article}

\usepackage{cvpr}              
\makeatletter
\let\LenToUnit\relax
\let\AtPageUpperLeft\relax

\let\AtTextUpperLeft\relax

\let\ESO@HookI\relax
\let\ESO@HookII\relax
\let\ESO@HookIII\relax
\let\AddToShipoutPicture\relax

\let\ESO@isMEMOIR\relax
\let\ESO@yoffsetI\relax
\let\ESO@yoffsetII\relax
\makeatother
\usepackage{pdfpages}

\usepackage{amsmath}
\usepackage{graphicx}
\usepackage{comment}
\usepackage{algorithm}
\usepackage{listings}
\usepackage{pythonhighlight}

\usepackage{array}
\usepackage{multirow}
\usepackage{tabularx}
\usepackage{extarrows}
\usepackage{booktabs}

\usepackage[resetlabels,labeled]{multibib}

\definecolor{cvprblue}{rgb}{0.21,0.49,0.74}
\usepackage[pagebackref,breaklinks,colorlinks,allcolors=cvprblue]{hyperref}

\def\paperID{} 
\def\confName{CVPR}
\def\confYear{2026}

\title{Improving Controllable Generation: Faster Training and Better Performance via $x_0$-Supervision}

\author{Amadou S. Sangare \and Adrien Maglo \and Mohamed Chaouch \and Bertrand Luvison\\
Universit\'{e} Paris-Saclay, CEA, List, F-91120, Palaiseau, France\\
{\tt\small \{amadou-siaka.sangare, adrien.maglo, mohamed.chaouch, bertrand.luvison\}@cea.fr}
}

\begin{document}
\twocolumn[{%
  \renewcommand\twocolumn[1][]{#1}%
  \maketitle
    \vspace{-12pt}
    \captionsetup{type=figure}
    \centering
    \includegraphics[width=0.93\textwidth]{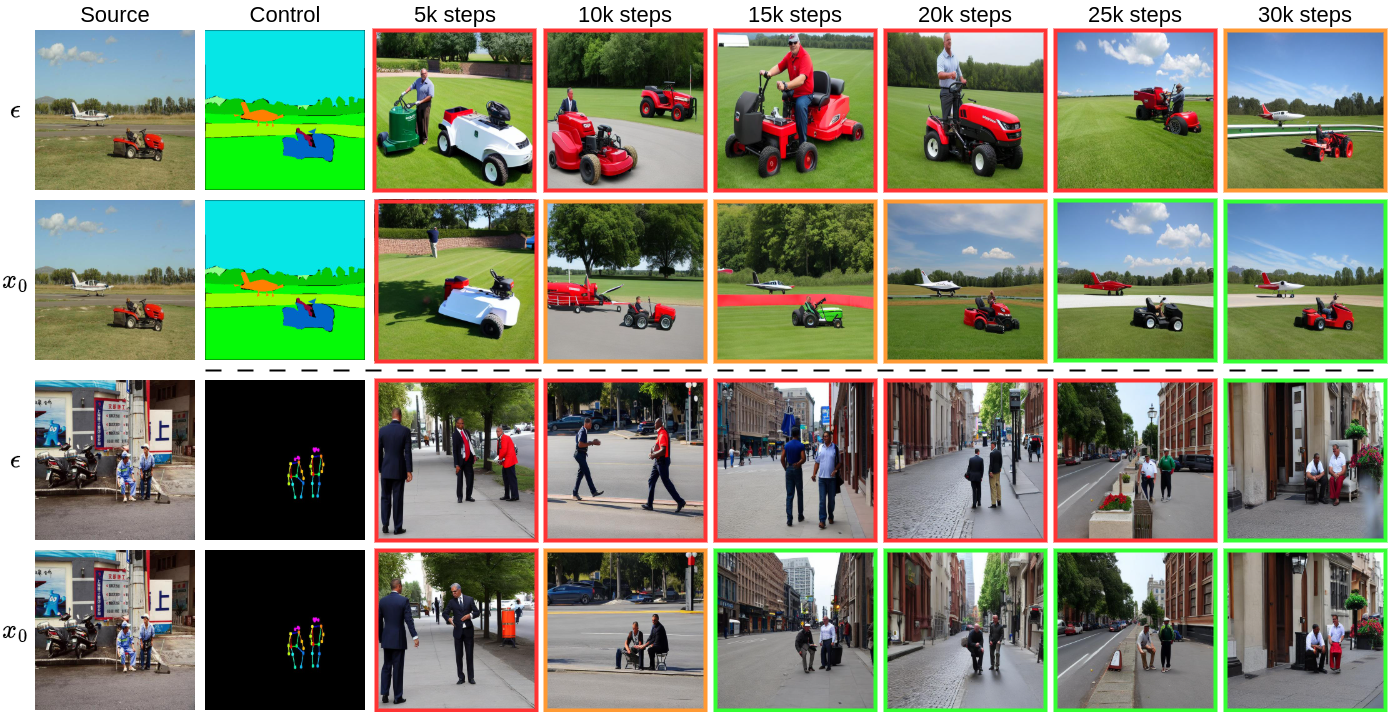}
    \vspace{-4pt}
    \caption{\textbf{ControlNet converges faster with the clean image $x_0$ as the supervision signal compared to the baseline $\epsilon$.} The red, orange, and green squares respectively indicate that the generated sample does not follow, partially follows, and correctly follows the input control.}
    \label{fig:teaser}
    \vspace{10pt}
}]
\begin{abstract}
Text-to-Image (T2I) diffusion/flow models have recently achieved remarkable progress in visual fidelity and text alignment. However, they remain limited when users need to precisely control image layouts, something that natural language alone cannot reliably express. Controllable generation methods augment the initial T2I model with additional conditions that more easily describe the scene. Prior works straightforwardly train the augmented network with the same loss as the initial network. Although natural at first glance, this can lead to very long training times in some cases before convergence. In this work, we revisit the training objective of controllable diffusion models through a detailed analysis of their denoising dynamics. We show that direct supervision on the clean target image, dubbed $x_0$-supervision, or an equivalent re-weighting of the diffusion loss, yields faster convergence. Experiments on multiple control settings demonstrate that our formulation accelerates convergence by up to 2$\times$ according to our novel metric (mean Area Under the Convergence Curve - mAUCC), while also improving both visual quality and conditioning accuracy. Our code is available at \url{https://github.com/CEA-LIST/x0-supervision}
\end{abstract}    
\section{Introduction}
\label{sec:intro}

Recent years have witnessed an unprecedented rise in generative AI, driven by breakthroughs in image generative modeling. Diffusion models~\cite{ho2020denoising,sohldickstein2015deepunsupervisedlearningusing,song2021scorebased,peebles2023DiT}, and more recently flow-based approaches~\cite{LipmanCBNL23,LiuG023,AlbergoV23}, now define the state of the art in visual fidelity and semantic alignment. However, this leap in quality has come at a substantial computational cost: training such models demands massive GPU memory, long optimization cycles, and considerable energy consumption. As generative AI continues to scale, the intensity of these resource requirements grows exponentially, raising both environmental and accessibility concerns. Developing more efficient, scalable, and cost-effective learning strategies has therefore become a key challenge for the next generation of generative models.

To bring more user-control to the sampling process, initial methods used natural language as the conditioning signal for the diffusion model. These models are referred to as Text-to-Image (T2I) models \cite{rombach2022highresolutionimagesynthesislatent,esser2024scalingrectifiedflowtransformers,blackforestlabs_flux}. Although successful in less constrained scenarios, their reliance on natural language alone limits their effectiveness in structured settings where the user might need very specific layouts that are difficult to express verbally. This limitation arises from the inherent ambiguity of natural language. This fundamental limitation has given rise to the field of controllable generation. The key idea is to augment a pre-trained T2I model with novel conditioning signals that more accurately capture the user’s intent. This is achieved by plugging an adapter network, taking in the novel conditioning signals (segmentation masks, edges, poses...), into the base model, and training the augmented model to follow them.

For clarity, we distinguish between the output predicted by the controllable generative network and the signal used to supervise that prediction. Specifically, \cite{zhang2023controlnet,mou2024t2i-adapter,zhao@2023uni-controlnet,qin2023unicontrol,li2023gligen,wang2024instancediffusion,zhou2024migc} plug an adapter into Stable Diffusion 1.4/1.5 \cite{rombach2022highresolutionimagesynthesislatent}, which is an $\epsilon$-predictor, and train the adapter with the same $\epsilon$-supervision loss. Similarly, \cite{tan2025ominicontrol} build on FLUX.1 \cite{blackforestlabs_flux}, which is a velocity $u$-predictor, and train with the original $u$-supervision loss. Although this is a natural and effective choice, it does not leverage the specific nature of controllable generation with spatial constraints (\textit{i.e.}, constraints on scene layout). Moreover, it is not optimal in terms of training speed.

A first consideration is the sampling process. 
As shown in \cref{fig:ddim_sampling}, images are generated in a coarse-to-fine manner during diffusion/flow sampling. In the first iterations, the overall layout is determined, and in the later iterations, fine-grained details are added while keeping the layout roughly fixed. In standard T2I generation, the first stage is not subject to strict constraints. In contrast, for spatial control, correctly determining the right layout is crucial, since errors made in this stage cannot be structurally corrected in the second stage.

A second consideration is how the network is trained to follow spatial controls. As mentioned earlier, it is desirable for the network to quickly learn to draw the right structure from spatial controls during the first denoising iterations. However, the naive choice of supervising the $\epsilon$-predictor with the $\epsilon$-supervision loss is equivalent to implicitly using the $x_0$-supervision loss \cite{salimans2022progressive} and applying a weighting function that severely penalizes the network to learn from these crucial early denoising iterations during training (see \cref{fig:snr_plot}). The same observation holds for flow matching.

Based on these observations, we propose to train any predictor using the $x_0$-supervision loss by appropriately converting the predicted signal into its equivalent $x_0$ estimate. Our contributions can be summarized as follows:

\begin{itemize}
    \item We propose a more efficient way to train controllable generative models. Our method is motivated by a careful and principled analysis of controllable image generation. The proposed $x_0$-supervision improves both training speed and final performance.
    \item Extensive experiments validate our approach and its underlying motivation. Our experimental results also provide an equivalent way to implement the method by suitably re-weighting the original supervision loss. Moreover, our experiments demonstrate that the proposed approach generalizes across paradigms (diffusion and flow matching) and architectures (UNet \cite{ronneberger2015UNet} and DiT \cite{peebles2023DiT}).
    \item We introduce a new metric, the \textbf{m}ean \textbf{A}rea \textbf{U}nder the \textbf{C}onvergence \textbf{C}urve (\textbf{mAUCC}), to measure convergence speed during controllable generation training. This metric is broadly applicable beyond this specific task and can be used to evaluate convergence in other training contexts.
\end{itemize}

\section{Related Works}
\label{sec:related_works}

\subsection{Image Generative models}
Generative modeling aims to approximate an unknown distribution by leveraging available samples from it. Successfully achieving this allows one to sample new instances from that distribution or estimate their likelihood. Significant advancements have been made over the past decade with the introduction of many paradigms. Variational Autoencoders (VAEs) \cite{kingma2022autoencodingvariationalbayes} train models based on an encoder-decoder architecture. The encoder maps the input to a latent space while the decoder maps the latent representation back to the original space. Generative Adversarial Networks (GANs) \cite{goodfellow2014generativeadversarialnetworks} learn to generate realistic samples by opposing a generator and a discriminator, the optimal state being when the discriminator can no longer distinguish real from generated samples. Continuous Normalizing Flows (CNFs) methods \cite{rezende2015variationalinferencenormalizingflows,chen2020residualflowsinvertiblegenerative} learn an invertible mapping between samples from a known distribution (e.g, a Gaussian distribution) to those of an unknown distribution of interest. Initial CNFs were trained via likelihood maximization, but this approach faces practical challenges such as costly ordinary differential equation (ODE) simulations and stringent architectural constraints. Flow Matching Models (FMMs) \cite{LipmanCBNL23,LiuG023,AlbergoV23} are trained to predict the velocity field driving the ODE that defines the CNF. This is a more efficient and tractable way to train CNFs. Diffusion Models (DMs) \cite{sohldickstein2015deepunsupervisedlearningusing,ho2020denoising,song2021scorebased} learn to reverse a Markovian diffusion process that gradually shifts samples from the original distribution to a target distribution, usually a Gaussian distribution in the literature. Flow Matching and Diffusion are equivalent in nature; both aim to map noise to clean data, although DMs are based on Stochastic Differential Equations (SDEs). FMMs and DMs establish the current state of the art.

\subsection{Controllable generation}
Text-to-image (T2I) \cite{rombach2022highresolutionimagesynthesislatent,esser2024scalingrectifiedflowtransformers,blackforestlabs_flux} generation has been the primary control signal used in the literature. T2I models, diffusion- or flow-based, leverage pre-trained language models like CLIP \cite{radford2021CLIP} or T5 \cite{2020t5} to encode text into latent representations and condition image generation using cross-attention \cite{rombach2022highresolutionimagesynthesislatent} or by concatenating the text and visual tokens \cite{esser2024scalingrectifiedflowtransformers}. 

Although achieving compelling results, these methods provide limited control when relying solely on textual descriptions, since natural language is limited in precisely describing the spatial composition of a scene.
Building on Stable Diffusion (SD) \cite{rombach2022highresolutionimagesynthesislatent}, ControlNet \cite{zhang2023controlnet} and T2I-Adapter \cite{mou2024t2i-adapter} first introduce adapters taking novel control signals. In the same line of work, OminiControl \cite{tan2025ominicontrol} introduces an efficient extension to the DiT \cite{peebles2023DiT} architecture. These types of method, however, require training a specific adapter for each control modality, which does not scale well with the number of control modality. To overcome this limitation, all-in-one methods \cite{zhao@2023uni-controlnet,qin2023unicontrol} train a single adapter capable of handling multiple control modalities. Instance-level \cite{li2023gligen,wang2024instancediffusion,zhou2024migc} methods increase the granularity of control by giving the ability to control each instance in the image with location information (\eg bounding box), shape information (\eg pose, segmentation) and appearance information such as textual description.

\subsection{Diffusion parameterizations}
Diffusion models can be trained with different objectives. Each objective gives a different parameterization of the diffusion. Three major parameterizations exist in the literature. During training, an image and random noise are linearly combined with coefficients given by a noising schedule that depends on the time step. Given the noisy image, the network either predicts the noise ($\epsilon$-prediction) \cite{ho2020denoising} or image ($x_0$-prediction) \cite{salimans2022progressive}. \citet{salimans2022progressive} introduced another objective, that is a combination of the two, called $v$-prediction enabling a more stable training.

In the image domain, most controllable generation methods \cite{zhang2023controlnet,mou2024t2i-adapter,zhao@2023uni-controlnet,qin2023unicontrol,li2023gligen,wang2024instancediffusion,zhou2024migc} are based on the versions of Stable Diffusion that have been initially trained with $\epsilon$-prediction. Hence, they proceed with controllable generation training using the same training loss, here $\epsilon$-prediction loss. However, this approach is not the most efficient and can lead to very long training times before convergence, depending on the controlled generation task. In this work we propose using the $x_0$-supervision loss as a more efficient way to train the controlled model.

\begin{figure*}[t!]
  \includegraphics[width=\linewidth]{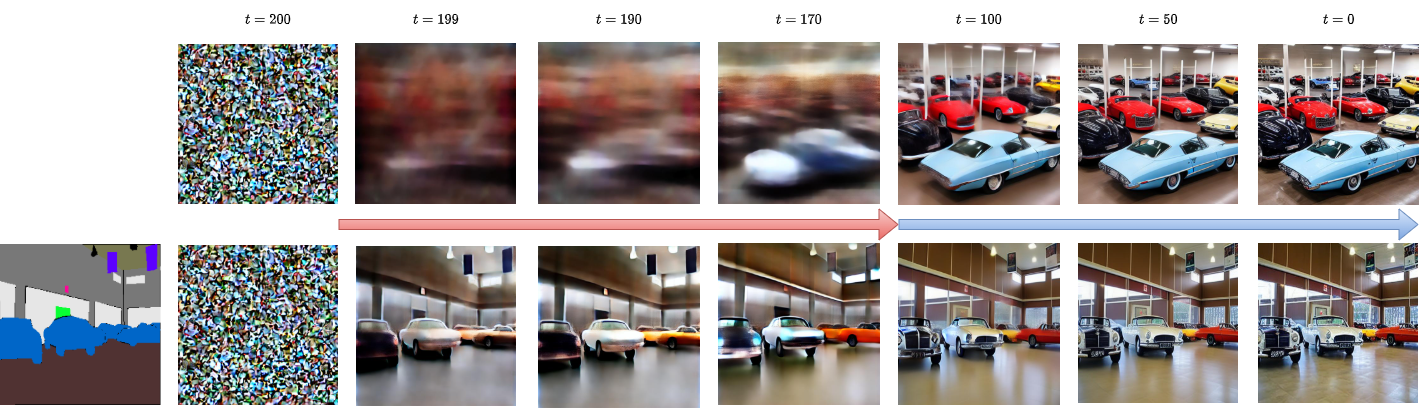}
  \caption{\textbf{Evolution of the final image during a DDIM \cite{song2021denoising} sampling with $200$ steps.} The top row is the result for Stable Diffusion $1.5$ and the bottom row is the result for a segmentation ControlNet \cite{zhang2023controlnet} based on Stable Diffusion $1.5$. As we can see, images are generated in a coarse-to-fine fashion. The early steps determine the overall layout of the scene (red arrow), while the next steps add fine-grained details (blue arrow). Since semantic segmentation already reveals the target layout, the coarse phase is much faster and the model have more time to add fine-grained details.}
  \label{fig:ddim_sampling}
\end{figure*}%

\begin{figure}[h!]
  \includegraphics[width=\linewidth]{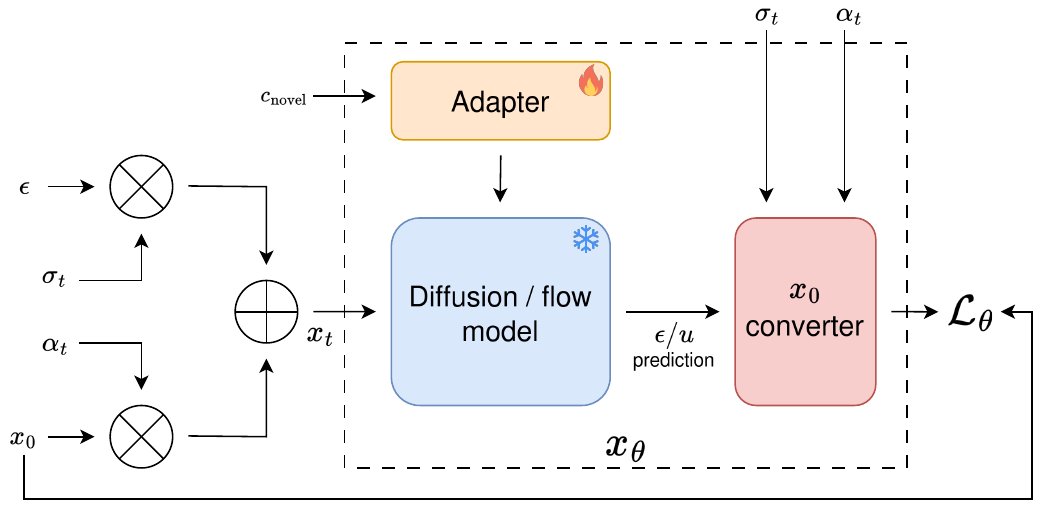}
  \caption{\textbf{Our approach.} For a more efficient controllable generation training, we propose to convert any predictor to an $x_0$-predictor, and supervise with the clean image. This simple trick significantly improves the convergence speed and the final performance.
  }
  \label{fig:any-to-x0}
\end{figure}%
\section{Approach}
\label{sec:approach}

\begin{figure*}[htbp]
    \centering
    
    \subfloat[ControlNet Depth. \label{fig:controlnet_depth_curve}]{
        \begin{minipage}{0.22\textwidth}
            \centering
            \includegraphics[width=\linewidth]{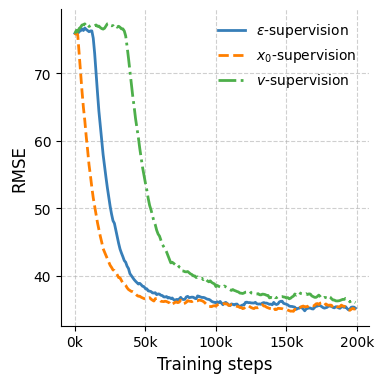}
            
        \end{minipage}
    }
    \subfloat[ControlNet Segmentation. \label{fig:controlnet_seg_curve}]{
        \begin{minipage}{0.22\textwidth}
            \centering
            \includegraphics[width=\linewidth]{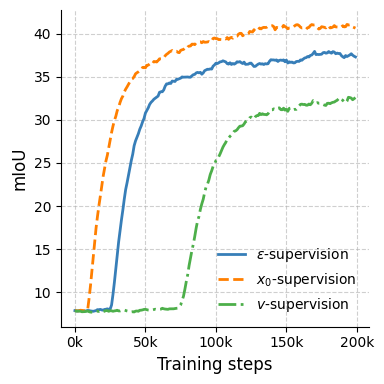}
        \end{minipage}
    }
    \subfloat[ControlNet Canny Edge. \label{fig:controlnet_canny_curve}]{
        \begin{minipage}{0.22\textwidth}
            \centering
            \includegraphics[width=\linewidth]{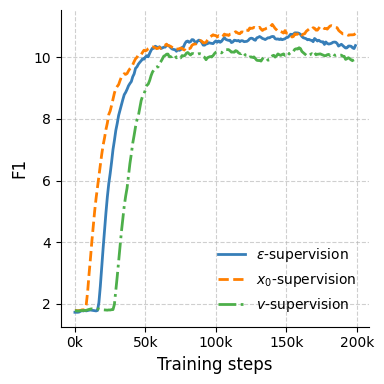}
        \end{minipage}
    }
    \subfloat[ControlNet Pose. \label{fig:controlnet_pose_curve}]{
        \begin{minipage}{0.22\textwidth}
            \centering
            \includegraphics[width=\linewidth]{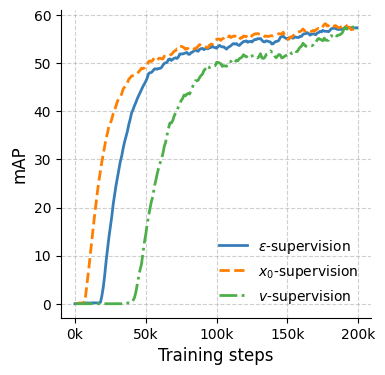}
        \end{minipage}
    }
    
    \vspace{0.2cm} 
    
    \subfloat[OminiControl Depth. \label{fig:omini_depth_curve}]{
        \begin{minipage}{0.22\textwidth}
            \centering
            \includegraphics[width=\linewidth]{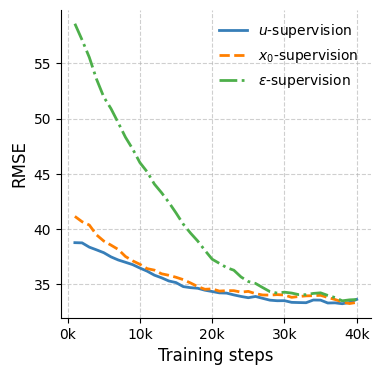}
            
        \end{minipage}
    }
    \subfloat[OminiControl Segmentation. \label{fig:omini_seg_curve}]{
        \begin{minipage}{0.22\textwidth}
            \centering
            \includegraphics[width=\linewidth]{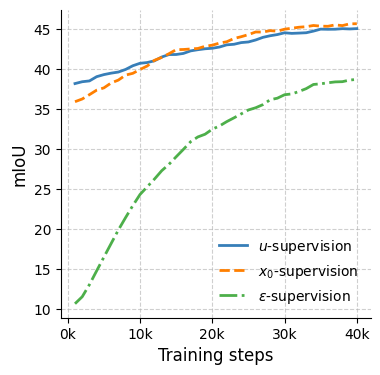}
        \end{minipage}
    }
    \subfloat[OminiControl Canny Edge. \label{fig:omini_canny_curve}]{
        \begin{minipage}{0.22\textwidth}
            \centering
            \includegraphics[width=\linewidth]{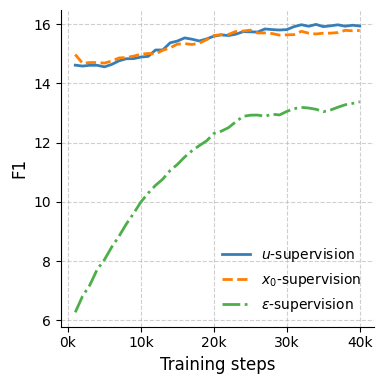}
        \end{minipage}
    }
    \subfloat[OminiControl Pose. \label{fig:omini_pose_curve}]{
        \begin{minipage}{0.22\textwidth}
            \centering
            \includegraphics[width=\linewidth]{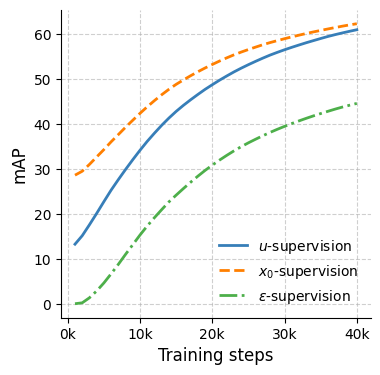}
        \end{minipage}
    }
    
    
%
    \caption{\textbf{Convergence curves for ControlNet and OminiControl on different tasks.} We use an EMA weight of $0.9$ to smooth the curves. We can notice that the convergence is faster with $x_0$-supervision.}
    \label{fig:main_convergence_curves}
\end{figure*}%

\begin{figure}[htbp]
    \centering
    
    \subfloat[GLIGEN Box+Text. \label{fig:gligen_box_text_curve}]{
        \begin{minipage}{0.45\columnwidth}
            \centering
            \includegraphics[width=0.95\linewidth]{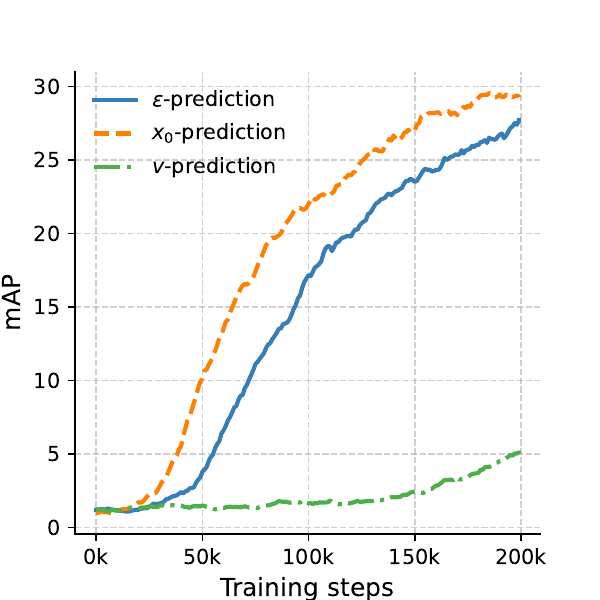}
            
        \end{minipage}
    }
    \subfloat[GLIGEN Box+Text+Image. \label{fig:gligen_box_text_image_curve}]{
        \begin{minipage}{0.45\columnwidth}
            \centering
            \includegraphics[width=0.95\linewidth]{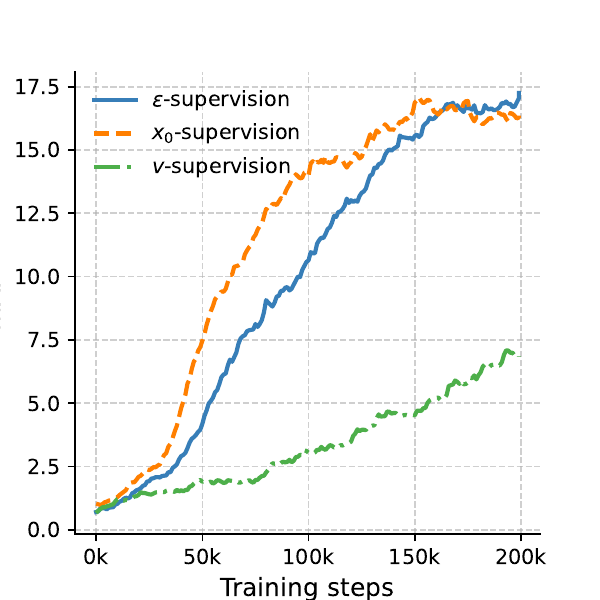}
        \end{minipage}
    }

    \caption{\textbf{Convergence curves for GLIGEN on different tasks.} We use an EMA weight of $0.9$ to smooth the curves. We can notice that the convergence is faster with $x_0$-supervision.}
    \label{fig:gligen_convergence_curves}
\end{figure}%

\subsection{Diffusion background}
\label{sec:diffusion_background}
Drawing inspiration from non-equilibrium thermodynamics, the diffusion paradigm aims to gradually shift data to noise thanks to a Markovian diffusion process, and train a denoising network to reverse this procedure. The diffusion process, also called the forward process, is predefined by a noise schedule $(\alpha_t, \sigma_t)$ where $t$ is the timestep that can be either discrete ($t\in\{0,\dots,T\}$) or continuous ($t\in[0, 1]$). Formally, given $x_0$ and timestep $t$:
\begin{equation}
    p(x_t|x_0) = \mathcal{N}(x_t;\alpha_t x_0,\sigma_t^2\mathbf{I}),
\end{equation}
where $x_t$ is the resulting noisy sample, also referred to as the interpolant, and can be expressed as $x_t = \alpha_t x_0 + \sigma_t \epsilon$ where $\epsilon\sim\mathcal{N}(0, \mathbf{I})$. Two major types of schedules exists in the literature, variance-preserving schedules and variance-exploding schedules \cite{song2021scorebased}. \citet{kingma2021vdm} showed that they are equivalent up to a rescaling of the noisy sample $x_t$. Our experiments are based on the versions of Stable Diffusion that uses the variance-preserving schedule in which $\alpha_t^2 + \sigma_t^2 = 1$.
A network with parameters $\theta$ is then trained to either predict $\epsilon$ \cite{ho2020denoising} ($\epsilon$-prediction):
\begin{equation}
    \mathcal{L}_{\theta}^{\epsilon} = \mathbb{E}_{t,\epsilon,x_0\sim p(x_0)}[\|\epsilon - \epsilon_{\theta}(x_t,t)\|_2^2],
\end{equation}
or predict $x_0$ \cite{salimans2022progressive} ($x_0$-prediction):
\begin{equation}
    \mathcal{L}_{\theta}^{x_0} = \mathbb{E}_{t,\epsilon,x_0\sim p(x_0)}[\|x_0 - x_{\theta}(x_t,t)\|_2^2],
\end{equation}
or another objective introduced by \cite{salimans2022progressive} called $v$-prediction:
\begin{equation}
    \mathcal{L}_{\theta}^{v} = \mathbb{E}_{t,\epsilon,x_0\sim p(x_0)}[\|v_t - v_{\theta}(x_t,t)\|_2^2],
\end{equation}
where $v_t = \alpha_t \epsilon - \sigma_t x_0$. Although these losses are equivalent up to a certain weighting, they have different behavior in terms of training stability and efficiency.

For sampling, many algorithms have been proposed in the literature, such as using a numerical method based on the Probability Flow ODE \cite{song2021scorebased}, or directly sampling from the posterior of the forward distribution as done with DDPM \cite{ho2020denoising} or DDIM \cite{song2021denoising}. The reverse process of DDPMs and DDIMs satisfy \cite{song2021denoising}:
\begin{equation}
    x_{t-1} = \alpha_{t-1}\underbrace{\left(\frac{x_t - \sigma_t\epsilon_{\theta}(x_t, t)}{\alpha_t}\right)}_{x_0\text{-prediction at step t}} + \sqrt{\sigma_{t-1}^2 - \gamma_t^2}\epsilon_{\theta}(x_t, t) + \gamma_t\epsilon_t,
\end{equation}
where $\epsilon_t\sim\mathcal{N}(0, \mathbf{I})$ is standard Gaussian noise re-injected at each denoising step during sampling and $\gamma_t$ controls the stochasticity of the sampling process. The case where $\gamma_t\equiv0$ corresponds to DDIM and the case where $\gamma_t = \frac{\sigma_{t-1}}{\sigma_t}\sqrt{1 - \frac{\alpha_t^2}{\alpha_{t-1}^2}}$ corresponds to DDPM. We can notice that sampling works by predicting $x_0$ based on $x_t$ and determining the corresponding $x_{t-1}$ using the forward process. This is repeated until $t=0$. On \cref{fig:ddim_sampling}, we can see that samples are generated in a coarse-to-fine manner. During the early steps (near $T$), the model determine the rough layout of the scene. It then adds fine-grained details in the following steps. For a T2I model, the first stage can take more time because of the uncertainty on the layout to chose. While in a controllable generation setting, the user already gives this layout, so the model focus more on adding details. Hence, the predicted $x_0$ at $t=199$ is quite blurry for Stable Diffusion but clear for the segmentation ControlNet. These observations mainly show that correctly predicting $x_0$ in the first stage is very important, with spatial control in particular, since it significantly determines the final results.

\subsection{Adding novel control signals}
\label{sec:adding_novel_controls}
Based on a Stable Diffusion trained on T2I generation with $\epsilon$-prediction:
\begin{equation}
    \mathcal{L}_{\theta}^{\epsilon} = \mathbb{E}_{t,\epsilon,x_0\sim p(x_0)}\left[\|\epsilon - \epsilon_{\theta}(x_t, c_{\text{text}}, t)\|_2^2\right],
\end{equation}
where $c_{\text{text}}$ is the text prompt, most controllable generation methods  \cite{zhang2023controlnet,mou2024t2i-adapter,zhao@2023uni-controlnet,qin2023unicontrol,li2023gligen,wang2024instancediffusion,zhou2024migc} add the novel control signals and directly continue training with $\epsilon$-supervision:
\begin{equation}
    \mathcal{L}_{\theta}^{\epsilon} = \mathbb{E}_{t,\epsilon,x_0\sim p(x_0)}\left[\|\epsilon - \epsilon_{\theta}(x_t, c_{\text{text}}, c_{\text{novel}}, t)\|_2^2\right],
\end{equation}
where $c_{\text{novel}}$ is the novel control signal. The Stable Diffusion backbone is frozen during training, only the adapter weights are updated.

Although this approach is a natural choice, we argue that this is sub-optimal in the case of controllable generation. This is particularly true for cases where the control signals are not explicitly spatially aligned with the target image. One example is the control method of GLIGEN \cite{li2023gligen} that uses instance-level conditions such as a (bounding box coordinates, text) pair to control the layout of the image. This type of method is considerably slower to converge compared to methods like ControlNet \cite{zhang2023controlnet} for which the spatial alignment between the control signals and the target image is explicit. Based on the discussion in \cref{sec:diffusion_background}, we propose to train by supervising with the clean data $x_0$. Intuitively, the strong correlation between the final image and the control signals reduces the complexity of the $x_0$-prediction task, even at very low signal-to-noise ratios (near $T$). Moreover, this forces the model to quickly learn to pick the right layout, at the beginning of denoising. Since Stable Diffusion 1.4/1.5 is an $\epsilon$-predictor, we use its predicted noise to derive an $x_0$-predictor:
\begin{equation}
    x_{\theta}(x_t, c_{\text{text}}, c_{\text{novel}}, t) = \frac{x_t - \sigma_t \epsilon_{\theta}(x_t, c_{\text{text}}, c_{\text{novel}}, t)}{\alpha_t},
\end{equation}
and train the model with:
\begin{equation}
    \mathcal{L}_{\theta}^{\epsilon\rightarrow x_0} = \mathbb{E}_{t,\epsilon,x_0\sim p(x_0)}\left[\|x_0 - x_{\theta}(x_t, c_{\text{text}}, c_{\text{novel}}, t)\|_2^2\right],
\end{equation}
\cref{fig:any-to-x0} is an illustration. This aligns with the network preconditioning proposed by \cite{Karras2022edm} and also used in the consistency training literature \cite{song2023consistency,luo2023latent} with:
\begin{equation}
    x_{\theta}(x_t, c_{\text{text}}, c_{\text{novel}}, t) = c_{\text{skip}}(t)\cdot x_t + c_{\text{out}}(t)\cdot\epsilon_{\theta}(x_t, c_{\text{text}}, c_{\text{novel}}, t),
\end{equation}
with $c_{\text{skip}}(t) = \frac{1}{\alpha_t}$, and $c_{\text{out}}(t) = -\frac{\sigma_t}{\alpha_t}$.

Since we keep the original objective ($\epsilon$ here) and suitably convert it before supervising with $x_0$, \textbf{$\mathbf{x_0}$-supervision} is a more appropriate term to describe our method.
Our experiments show its improvements over to the widespread approach of continuing with the $\epsilon$-supervision loss. One should notice that our approach does not apply if the base model is already an $x_0$-predictor.

\subsection{Formal justification}
\label{sec:method_justification}
\begin{figure}[h!]
    \centering
    
    \subfloat[Noise schedule. \label{fig:sd_noise_schedule}]{
        \begin{minipage}{0.45\columnwidth}
            \centering
            \includegraphics[width=\textwidth]{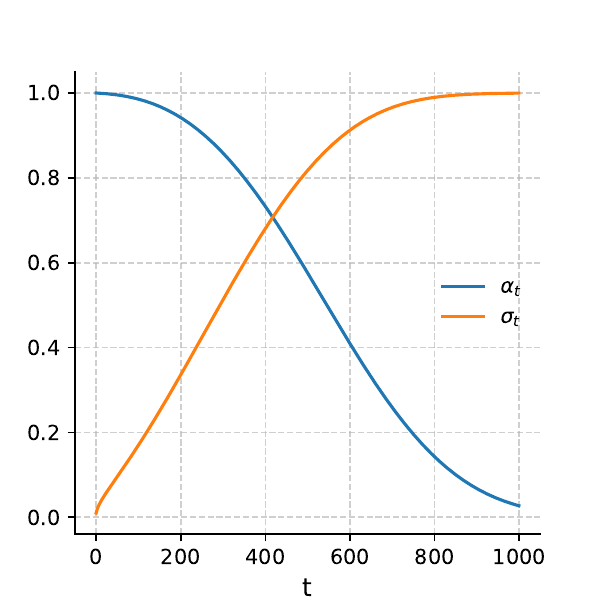}
        \end{minipage}
    }
    \subfloat[Signal-to-Noise ratio. \label{fig:sd_snr}]{
        \begin{minipage}{0.45\columnwidth}
            \centering
            \includegraphics[width=\textwidth]{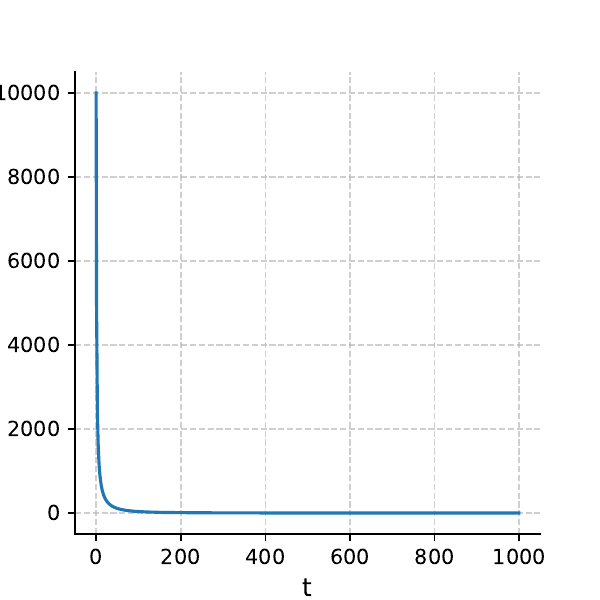}
        \end{minipage}
    }
    
    \caption{\textbf{The noise schedule and the evolution of the signal-to-noise ratio in Stable Diffusion.} The SNR decrease very quickly, using it as the loss weighting function significantly down-weights the learning signal for low SNRs.}
    \label{fig:snr_plot}
\end{figure}

As explained by \cite{salimans2022progressive}, an $\epsilon$-predictor is an implicit $x_0$-predictor. Temporarily omitting the expectations for notation convenience, the loss can be derived as follows:
\begin{align}
    \mathcal{L}_{\theta}^{\epsilon} &= \|\epsilon - \epsilon_{\theta}(x_t, c_{\text{text}}, c_{\text{novel}}, t)\|_2^2, \\
    &= \|\frac{1}{\sigma_t}(x_t - \alpha_t x_0) - \frac{1}{\sigma_t}(x_t - \alpha_t x_{\theta}(x_t, c_{\text{text}}, c_{\text{novel}}, t))\|_2^2,  \\
    &= \frac{\alpha_t^2}{\sigma_t^2}\|x_0 - x_{\theta}(x_t, c_{\text{text}}, c_{\text{novel}}, t)\|_2^2 = \frac{\alpha_t^2}{\sigma_t^2} \mathcal{L}_{\theta}^{x_0},
\end{align}
therefore, the $\epsilon$-supervision loss is equivalent to weighting the $x_0$-supervision loss with the signal-to-noise ratio $\frac{\alpha_t^2}{\sigma_t^2}$ (SNR). As it can be noticed on \cref{fig:snr_plot}, the SNR is very high in late denoising steps (near $0$) and then goes to zero very quickly. Hence the loss is given near-zero weights for low SNRs. This results in the model having very low learning signal for early denoising steps (near $T$). But our discussion in \cref{sec:diffusion_background} shows that these steps determine the global composition of the image. Deriving our loss gives:
\begin{align}
    \mathcal{L}_{\theta}^{\epsilon\rightarrow x_0} &= \|x_0 - x_{\theta}(x_t, c_{\text{text}}, c_{\text{novel}}, t)\|_2^2, \\
    &= \|\frac{1}{\alpha_t}(x_t - \sigma_t\epsilon) - x_{\theta}(x_t, c_{\text{text}}, c_{\text{novel}}, t)\|_2^2, \\
    &= \frac{\sigma_t^2}{\alpha_t^2}\|\epsilon - \epsilon_{\theta}(x_t, c_{\text{text}}, c_{\text{novel}}, t)\|_2^2, \\
    &= \frac{\sigma_t^2}{\alpha_t^2}\mathcal{L}_{\theta}^{\epsilon} = \mathcal{L}_{\theta}^{x_0},
\end{align}
Therefore, we can see that our explicit $x_0$-supervision removes the weighting, allowing the model to a get stronger learning signals from early denoising steps, besides, since the control signals and the target image are strongly correlated, the denoising task at these early steps becomes easier compared to the T2I case. We empirically validate this on ControlNet by training with the following loss:
\begin{equation}
    \mathcal{L}_{\theta}' = \mathbb{E}_{t,\epsilon,x_0\sim p(x_0)}\left[\frac{\sigma_t^2}{\alpha_t^2}\|\epsilon - \epsilon_{\theta}(x_t, c_{\text{text}}, c_{\text{novel}}, t)\|_2^2\right], \label{eq:inv_snr_eps_loss}
\end{equation}
which is simply weighting the original $\epsilon$-supervision loss with the inverse of the SNR.

\subsection{Mean Area Under the Convergence Curve}

Taking inspiration from the active learning literature \cite{guyonCDL10,viering2023learningcurves}, we provide a new metric, called \textbf{mAUCC} to measure convergence speed. Given the convergence curve at the end of training, we compute the average area under this curve over various training horizons. This makes \textbf{mAUCC} less sensitive to the training horizon compared to the simple area under the curve (see Supp. A.2). Formally:
\begin{align}
    \textbf{\text{AUCC}}@t_i &= \frac{1}{\lceil t_i T_{max}\rceil} \int_{0}^{\lceil t_i T_{max}\rceil}m_s d_s \\
    \textbf{\text{mAUCC}} &= \frac{1}{N_{th}}\sum_{i=1}^{N_{th}}\textbf{\text{AUCC}}@t_i
\end{align}
where $m_s$ is the normalized metric at step $s$, $N_{th}$ is the number training time horizons, $\textbf{\text{AUCC}}@t_i$ is the area under the convergence curve for a horizon of $t_i$ percent of the maximum training iterations $T_{max}$. In our experiments, we use percentages from $25\%$ to $100\%$ with a step of $5\%$. Intuitively, one \textbf{AUCC} measures the cumulative performance over training iterations. It measures how fast the curve increases or decreases. \textbf{mAUCC} is the average of the \textbf{AUCC} over various training horizons. If the underlying metric is a score, then higher is better. In contrast, if it is an error or a distance, then lower is better. The metrics and the number of iterations are normalized to $[0, 1]$ in order to have a normalized \textbf{mAUCC}.

\section{Experiments}
\label{sec:experiments}

\begin{table*}[htbp]
  \centering
  \vspace{-10pt}
  \resizebox{\textwidth}{!}{%
    \begin{tabular}{c|ccc|ccc|ccc|ccc}
    \toprule
    \multirow{2}[2]{*}{Supervision} & \multicolumn{3}{c|}{Depth} & \multicolumn{3}{c|}{Semantic Seg} & \multicolumn{3}{c|}{Canny Edge} &  \multicolumn{3}{c}{Pose} \\
          & FID$\downarrow$ & RMSE$\downarrow$ & mAUCC$\downarrow$  & FID$\downarrow$ & mIoU$\uparrow$ & mAUCC$\uparrow$ & FID$\downarrow$ & F1$\uparrow$ & mAUCC$\uparrow$ & FID$\downarrow$ & mAP$^{\text{pose}}\uparrow$ & mAUCC$\uparrow$ \\
    \midrule
    \multicolumn{13}{c}{ControlNet (Diffusion)} \\
    \midrule
    $\epsilon$-ControlNet & $17.68$ & $35.79$ & $17.70$ & $30.05$ & $35.84$ & $25.19$ & $31.18$ & $9.22$ & $8.09$ & $44.09$ & $58.00$  & $35.86$\\
    $v$-ControlNet & $17.82$ & $36.53$ & $21.89$ & $32.17$ & $31.11$ & $13.48$ & $31.61$ & $8.92$ & $6.93$ & $44.32$ & $52.34$ & $22.72$ \\
    $x_0$-ControlNet & $\textbf{17.50}$ & $\textbf{35.42}$ & $\textbf{15.98}$ & $\textbf{29.55}$ & $\textbf{39.54}$ & $\textbf{31.52}$ & $\textbf{30.31}$ & $\textbf{9.56}$ & $\textbf{8.76}$ & $\textbf{44.09}$ & $\textbf{59.18}$ & $\textbf{42.19}$ \\
    \midrule
    \multicolumn{13}{c}{T2I-Adapter (Diffusion)} \\
    \midrule
    $\epsilon$-T2I-Adpater & $20.90$ & $49.48$  & $22.44$ & $33.72$ & $23.48$ & $17.39$ & $34.38$ & $4.10$ & $3.61$ & $\textbf{46.76}$ & $29.50$ & $12.98$ \\
    $v$-T2I-Adapter & $23.09$ & $61.43$ & $26.52$ & $37.87$ & $15.09$ & $11.54$ & $38.57$ & $2.89$ & $2.62$ & $48.91$ & $11.93$ & $4.44$ \\
    $x_0$-T2I-Adapter & $\textbf{18.92}$ & $\textbf{42.35}$ & $\textbf{19.67}$ & $\textbf{32.90}$ & $\textbf{26.21}$ & $\textbf{21.07}$ & $\textbf{32.53}$ & $\textbf{4.89}$ & $\textbf{4.47}$ & $46.81$ & $\textbf{37.75}$ & $\textbf{21.45}$ \\
    \midrule
    \multicolumn{13}{c}{OminiControl (Flow Matching)} \\
    \midrule
    $u$-OminiControl & $18.48$ & $\textbf{34.97}$ & $\textbf{13.47}$ & $28.20$ & $42.09$ & $\textbf{39.13}$ & $27.53$ & $13.90$ & $14.40$ & $\textbf{36.93}$ & $\textbf{66.53}$ & $34.54$ \\
    $\epsilon$-OminiControl & $\textbf{17.49}$ & $35.92$ & $16.79$ & $\textbf{27.44}$ & $37.53$ & $23.76$ & $27.87$ & $11.58$ & $9.71$ & $38.76$ & $54.66$ & $17.57$ \\
    $x_0$-OminiControl & $18.98$ & $35.83$ & $13.74$ & $29.17$ & $\textbf{45.60}$ & $38.69$ & $\textbf{27.12}$ & $\textbf{13.96}$ & $\textbf{14.41}$ & $39.19$ & $64.72$ & $\textbf{42.07}$ \\
    \bottomrule
    \end{tabular}%
    }
    \caption{\textbf{Comparison between the different supervision signals on different spatially-aligned control modalities and methods.} The curves are smoothed with EMA weight $0.9$ before computing the \textbf{mAUCC}.}
  \label{tab:main_results_spatially_aligned}%
\end{table*}

\begin{table}[htbp]
  \centering
  \vspace{-10pt}
  \resizebox{\columnwidth}{!}{%
    \begin{tabular}{c|ccc|ccc}
    \toprule
    \multirow{2}[2]{*}{Supervision} & \multicolumn{3}{c|}{Box+Text} & \multicolumn{3}{c}{Box+Text+Image} \\
          & FID$\downarrow$ & mAP$^{\text{box}}\uparrow$ & mAUCC$\uparrow$ & FID$\downarrow$ & mAP$^{\text{box}}\uparrow$ & mAUCC$\uparrow$  \\
    \midrule
    \multicolumn{7}{c}{GLIGEN (Diffusion)} \\
    \midrule
    $\epsilon$-GLIGEN & $32.58$ & $30.70$ & $8.28$ & $\textbf{21.40}$ & $\textbf{21.31}$ & $6.15$ \\
    $v$-GLIGEN & $35.99$ & $8.39$ & $1.54$ & $23.56$ & $2.18$ & $9.56$ \\
    $x_0$-GLIGEN & $\textbf{28.38}$ & $\textbf{33.30}$ & $\textbf{18.38}$ & $24.23$ & $20.76$ & $\textbf{8.07}$ \\
    \bottomrule
    \end{tabular}%
    }
    \caption{\textbf{Comparison between the different supervision signals on different non-spatially-aligned control modalities.} The curves are smoothed with EMA weight $0.9$ before computing the \textbf{mAUCC}.}
  \label{tab:main_results_non_spatially_aligned}%
\end{table}

\subsection{Methodology}

\textbf{Methods.} We restrict our experiments to few methods for reasons described hereafter. The control signals used in the current literature can be organised in two families. \textbf{Spatially-aligned} controls are those represented as images (\eg, \textit{depth map, segmentation mask, Canny edges, etc.}). For this type of control, the relation to the target image is explicit to the model, and convergence is usually faster with them. \textbf{Non-spatially-aligned} controls are those not represented to the model in image-form (\eg, \textit{bounding box coordinates, style or subject tokens, etc.}). For this type of control, since the relation to the target image is implicit, the model takes more time to learn it, leading to a longer convergence time. For \textbf{spatially-aligned} controls, we use ControlNet \cite{zhang2023controlnet} and T2I-Adapter \cite{mou2024t2i-adapter} as representatives, while for \textbf{non-spatially-aligned} controls, we use GLIGEN \cite{li2023gligen} as the representative method. We also include OminiControl \cite{tan2025ominicontrol} to evaluate our approach under the flow-matching paradigm. These choices are also motivated by the availability of their training codes.\smallskip

\noindent\textbf{Training setting.} For each method, we use the same training hyperparameters and the same types of GPUs to compare the different supervision signals during both training and evaluation. We use $40$k iterations  for OminiControl and $200$k training iterations for the other methods.
For diffusion, we experiment with the $\epsilon$-, $v$-, and $x_0$-supervision losses, and for flow matching, we experiment with the $\epsilon$-, $u$-, and $x_0$-supervision losses. In the result tables, the methods are prefixed by the type of supervision used. We use batch size $8$ for T2I-Adapter and OminiControl, and $64$ for GLIGEN following the original works. Regarding ControlNet, the authors used different batch sizes for different tasks, but we found that a batch size of $8$ works fine for the studied tasks. We use NVIDIA A100 (80GB), H100 (94GB), and H200 (141GB) GPUs.\smallskip

\noindent\textbf{Datasets.}
We use the MultiGen-20M \cite{li2020controlnet_plus_plus} dataset for depth control, ADE20K \cite{zhou2017ade20k} for semantic segmentation and Canny-edge control, and MS-COCO \cite{lin2014COCO} for pose control. We apply the Canny-edge detector to the semantic segmentations in order to restrict edges to object contours. Inner object edges are highly sensitive to the chosen thresholds \cite{sullivan2024edge_detection}. This prevents a consistent evaluation. Restricting to the contour edges allows a robust evaluation.  For GLIGEN \cite{li2023gligen}, we use the authors-provided datasets and evaluate the method on the MS-COCO subset from \cite{wang2024instancediffusion}.\smallskip

\noindent\textbf{Metrics.}
For visual quality assessment, we compute the Fr\'echet Inception Distance (FID) \cite{heusel2017ganstrainedtimescaleupdate} between generated images and real validation images.
To evaluate the fidelity of a model to a given control modality, we first estimate that modality on the generated images and then compare the estimated control with the input control. For depth, we use Midas \cite{ranftl2022midas} as the estimator and the Root Mean Squared Error (RMSE) as the evaluation metric. For semantic segmentation, we use MaskFormer Swin-L \cite{cheng2021maskformer} trained on ADE20K as the estimator and mIoU as the evaluation metric. For Canny-edge evaluation, we chain MaskFormer and the Canny edge detector. We use $100$ and $200$ as the upper and lower threshold following \cite{li2020controlnet_plus_plus}, and use the F1-score as the metric, again following \cite{li2020controlnet_plus_plus}. For pose, we use YOLO-Pose-11m \cite{yolo11_ultralytics} and MS-COCO's official evaluation tool to compute the $\text{mAP}^{\text{pose}}$. For GLIGEN's box modalities, we use YOLO-Det-11m \cite{yolo11_ultralytics} as the estimator and MS-COCO's evaluation tool to compute the $\text{mAP}^{\text{box}}$. All of these scores are computed on the individual images before averaging them. This better reflects how well the models follow the controls at inference. The final results reported on \cref{tab:main_results_spatially_aligned,tab:main_results_non_spatially_aligned} are all computed on the full validation sets.  To draw the convergence curves on \cref{fig:main_convergence_curves,fig:gligen_convergence_curves,fig:controlnet_x0_vs_inv_snr_convergence_curves}, these scores are evaluated after each $1$k iterations on a fixed set of $32$ images from the corresponding validation sets for OminiControl, and $64$ images for the other methods. From these curves, we compute our proposed \textbf{mAUCC} metric to evaluate the convergence speeds (\textbf{RMSE}, \textbf{mIoU}, \textbf{mAP}, \textbf{F1}).

\begin{table}[htbp]
  \centering
  \resizebox{\columnwidth}{!}{%
    \begin{tabular}{c|ccc|ccc}
    \toprule
    \multirow{2}[2]{*}{Supervision} & \multicolumn{3}{c|}{Box+Text (mAUCC)} & \multicolumn{3}{c}{Box+Text+Image (mAUCC)} \\
          & 16 & 32 & 64 & 16 & 32 & 64 \\
    \midrule
    $\epsilon$-GLIGEN & $1.41$ & $2.58$  & $8.28$ & $1.77$ & $3.26$ & $6.15$ \\
    $x_0$-GLIGEN & $\textbf{1.71}$ & $\textbf{7.72}$  & $\textbf{18.38}$ & $\textbf{1.82}$ & $\textbf{4.50}$ & $\textbf{8.07}$  \\
    \bottomrule
    \end{tabular}%
    }
    \vspace{-2pt}
    \caption{\textbf{$x_0$-supervision has higher batch size efficiency.} Since non-spatially-aligned controls require high batch sizes to converge, we compare how reducing the batch size affects the convergence speed for $x_0$- and $\epsilon$-supervision.}
  \label{tab:gligen_batch_size_x0_vs_eps}%
\end{table}

\subsection{Results}

In all our experiments, the $v$-supervision for diffusion and $\epsilon$-supervision for flow matching underperform compared to the original supervisions of the tested approaches. This discussed in Supp. D. This section focuses on the quantitative results of our proposed $x_0$-supervision. Qualitative results are deferred to Supp. E.\smallskip

\noindent\textbf{Spatially-aligned control signals.}
\Cref{tab:main_results_spatially_aligned} shows the main results for spatially-aligned control signals. We observe that $x_0$-supervision consistently improves ControlNet and T2I-Adapter metrics for all tasks. In particular, for ControlNet, mAUCC is significantly improved on semantic and pose control by $25\%$ and $17.65\%$, respectively. Interestingly, the so-called \textit{sudden-convergence phenomenon} \cite{zhang2023controlnet} is observed earlier with $x_0$-supervision in all cases, as shown in \cref{fig:controlnet_depth_curve,fig:controlnet_seg_curve,fig:controlnet_canny_curve,fig:controlnet_pose_curve} and \cref{fig:teaser}. As for T2I-Adapter, a drastic improvement of $65.25\%$ in mAUCC is observed for pose control. Moreover, the final control fidelity is improved by $16.84\%$, $11.63\%$ and $27.97\%$ for depth, segmentation, and pose control, respectively. Regarding OminiControl, the convergence speed remains mostly similar to the baseline $u$-supervision on depth, semantic, and Canny edge control, with variations of $2\%$, $1.12\%$ and $0.07\%$, respectively.
This behavior can be attributed to the already rapid convergence of the baseline in these tasks,
making the contribution of our method not noticeable within the chosen observation horizon. This faster convergence is most likely due to the stronger FLUX.1 base model. However, a $21.80\%$ improvement in mAUCC is observed for pose control, showing that $x_0$-supervision consistently enhances convergence speed while maintaining at least equivalent performance in the worst case. Moreover, an $8.34\%$ improvement in mIoU is observed for semantic control.\smallskip

\begin{figure}[hbp]
    \centering
    
    \subfloat[ControlNet Depth. \label{fig:controlnet_x0_vs_inv_snr_depth}]{
        \begin{minipage}{0.45\columnwidth}
            \centering
            \includegraphics[width=0.95\linewidth]{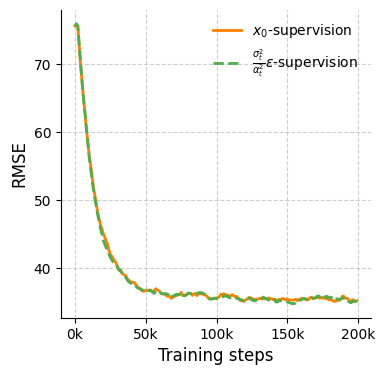}
            
        \end{minipage}
    }
    \subfloat[ControlNet Segmentation. \label{fig:controlnet_x0_vs_inv_snr_seg}]{
        \begin{minipage}{0.45\columnwidth}
            \centering
            \includegraphics[width=0.95\linewidth]{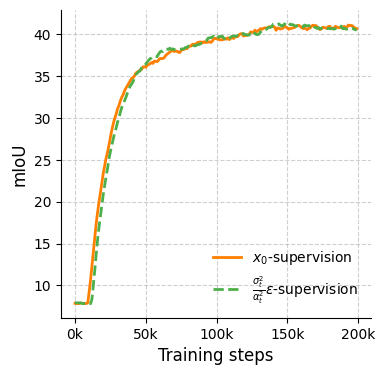}
        \end{minipage}
    }
    
    
    \subfloat[ControlNet Canny Edge. \label{fig:controlnet_x0_vs_inv_snr_canny}]{
        \begin{minipage}{0.45\columnwidth}
            \centering
            \includegraphics[width=0.95\linewidth]{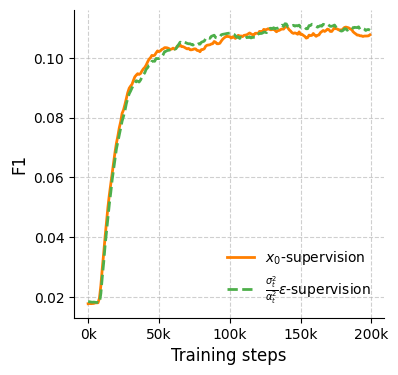}
            
        \end{minipage}
    }
    \subfloat[ControlNet Pose. \label{fig:controlnet_x0_vs_inv_snr_pose}]{
        \begin{minipage}{0.45\columnwidth}
            \centering
            \includegraphics[width=0.95\linewidth]{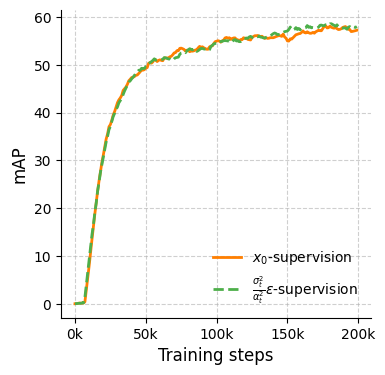}
        \end{minipage}
    }

    \caption{\textbf{Convergence comparison between $x_0$-supervision and $\frac{\sigma_t^2}{\alpha_t^2}\epsilon$-supervision.} We can see that they have the same convergence speed, hence validating our insights in \cref{sec:method_justification}.}
    \label{fig:controlnet_x0_vs_inv_snr_convergence_curves}
\end{figure}%

\noindent\textbf{Non-spatially-aligned control signals.}
\Cref{tab:main_results_non_spatially_aligned} shows the main results for non-spatially-aligned control signals on GLIGEN. The mAUCC is drastically improved by $121.98\%$ for box+text grounding and by $31.22\%$ for box+text+image grounding. Additionally, we improve mAP by $8.47\%$ for box+text grounding and maintain a similar performance for text+box+image, with a $2.58\%$ variation. Thus, the convergence speed improvement can be quite pronounced for these control types, since they tend to \textbf{converge slowly} and \textbf{require significant resources}.\smallskip

\noindent\textbf{Empirical validation of the formal justification.}
To empirically validate our justification in \Cref{sec:method_justification}, we trained ControlNet on all tasks using $\mathcal{L}_{\theta}'$ (see \cref{eq:inv_snr_eps_loss}). \Cref{fig:controlnet_x0_vs_inv_snr_convergence_curves} confirms that removing the SNR weighting yields a convergence speed similar to that of the $x_0$-supervision loss. This approach also provides an equivalent implementation of our method. The conversion formulas for all parameterizations and paradigms are exhaustively detailed in Supp. B.\smallskip

\noindent\textbf{Batch size efficiency of $x_0$-supervision on non-spatially-aligned controls.}
As discussed in \Cref{sec:adding_novel_controls}, since non-spatially-aligned controls are not given in image-form, the link to their corresponding region on the target image is not explicit. This results in higher requirements for batch size and training time, since the model takes more time to understand this link. For instance, \citet{li2023gligen} trained GLIGEN with a batch size of $64$ and $400$k training iterations. Training with a batch size of $64$ requires approximately $320$GB of VRAM (\eg, 4 NVIDIA A100 80GB GPUs), whereas training with a batch size of $8$ on a single NVIDIA A100 80GB works well for spatially-aligned controls.\smallskip

\noindent As shown in \Cref{tab:gligen_batch_size_x0_vs_eps}, $x_0$-supervision consistently leads to faster convergence across all batch sizes for GLIGEN. Noticeably, GLIGEN poorly learned the task after $200$k iterations with a batch size of $16$, further showing its large batch size requirements. Besides, comparable mAUCC values were achieved with a batch size of $32$ under $x_0$-supervision versus $64$ under $\epsilon$-supervision for box+text grounding, effectively reducing the memory requirements by half. Furthermore, increasing the batch size from $16$ to $64$ yields mAUCC improvements of $487.23\%$ for $\epsilon$-supervision and $974.85\%$ for $x_0$-supervision on box+text grounding. These findings demonstrate that $x_0$-supervision not only accelerates convergence but also mitigates memory constraints and scales more efficiently with batch size.
\section{Conclusion}
\label{sec:conclusion}

To partially address the major challenge of the soaring computational demands of generative AI models and the resulting environmental and accessibility concerns, we propose a simple yet highly effective method for training controllable image generative models.
Our method is motivated by a careful and principled analysis of the training and sampling of diffusion/flow models. Extensive experiments demonstrate that our method consistently improves convergence speed and final performance, achieving significant margins in some cases. We further demonstrate that our approach generalizes across paradigms (diffusion and flow matching) and architectures (UNets and DiTs). Moreover, our experiments show that our method also reduces resource requirements during training. Exploring further improvements is of major importance.

\section*{Acknowledgements}
This project was made possible by the
use of the FactoryIA supercomputer, financially supported by the Ile-De-France Regional Council.
This project was also provided with computing HPC and storage resources by GENCI at IDRIS thanks to the grant 2025-AD011016828 on the supercomputer Jean Zay's A100 and H100 partitions.
This work was supported in part by the European Union through the European Defence Fund (EDF) Project FaRADAI under Grant 101103386. Views and opinions expressed are those of the author(s) only and do not necessarily reflect those of the European Union nor the European Commission. Neither the European Union nor the granting authority can be held responsible for them.

{
    \small
    \bibliographystyle{ieeenat_fullname}
    \bibliography{main}
}


\clearpage
\onecolumn


\section*{Supplementary material}

# Improving Controllable Generation: Faster Training and Better Performance via $x_0$-Supervision

## Supplementary Material

Sec. A provides additional implementation details regarding our experiments and the computation of the mAUCC metric. Sec. B describes how to analytically convert the various diffusion and flow-matching supervision signals into an $x_0$ estimate. Sec. C comments on the convergence curves for T2I-Adapter. Sec. D offers insights into the suboptimal performance of the original $v$-ControlNet and $\epsilon$-OminiControl approaches. Finally, in Sec. E, we present further qualitative results illustrating the performance of our $x_0$-supervision for controllable generation.

## A. Further experimental details

### A.1. Controllable Generation Tasks

We evaluated our proposed $x_0$-supervision for controllable generation on several control modalities and previous frameworks so as to assess its effectiveness.

All training hyperparameters are set to the same values as in the original papers, with the exception of the ControlNet batch size, which is set to 8 in contrast to 32 for depth, 256 for semantic segmentation, 32 for Canny edge, and 18 for human pose in the original work. We generate images using the DDIM algorithm with 50 sampling steps and a guidance scale of 7.5. The same configuration is used for T2I-Adapter. For OminiControl, we use 28 sampling steps with a guidance scale of 3.5.

**Depth.** MultiGen-20M is used as the depth control dataset. We train the selected methods using the different supervision signals. Experiments are performed with images at a resolution of $512 \times 512$.

**Semantic Segmentation.** The ADE20K [20] dataset is used as the segmentation control dataset. We use Florence-2 [18] to generate image captions. Experiments are performed with images at a resolution of $512 \times 512$.

**Canny Edge.** We use the ADE20K dataset, annotated with Florence-2. In order to extract the control images, we apply the Canny edge detector to the semantic segmentation images to focus on object boundary edges. The aim is to ensure a robust evaluation protocol, as inner edges are highly sensitive to hysteresis thresholds.

**Pose.** We utilize the MS-COCO keypoints dataset [6]. We resize the images to a resolution of $512 \times 512$ and adjust the keypoint and bounding box coordinates accordingly. We retain only images where the bounding box covers at least $2\%$ of the total image area and which contain between 1 and 6 visible persons. Additionally, for each image, we use the first of the 5 captions provided in MS-COCO. We follow the same training and sampling procedures as described above.

**Box+Text grounding.** We utilize the dataset provided by the authors of GLIGEN [5]. It comprises a combination of GoldG [3], SBU [10], CC3M [14], and Object365 [13]. Further details can be found in the original paper. For training, we perform 200k iterations with a batch size of 64. For evaluation, we use the MS-COCO subset from [17]. The template "a <object>" is used for object captions, where <object> is replaced by the actual class name of the object. These captions are further embedded using the CLIP text encoder, as in the original paper. We generate images using the PLMS sampler [8] with 50 sampling steps and a guidance scale of 7.5, following the procedure described in the original paper. YOLO detection scores are computed image-wise and then averaged.

**Box+Text+Image grounding.** We use the same dataset as described above, following the original paper. We train for 200k iterations with a batch size of 64. The same MS-COCO subset is used for evaluation. In addition to object caption embeddings, object image embeddings are computed following the same procedure as in the original work. Finally, the same hyperparameters are used for sampling.

### A.2. mAUCC definition and implementation

In this section, we provide additional justifications to the definition of **mAUCC**. The area under the control fidelity curves is a reasonable way to assess the convergence speed since it faithfully represents the cumulative performance gains for a given training horizon. One natural choice is then to report $\mathbf{AUCC}@T_{max}$, where $T_{max}$ is the number of steps used for training. However, this is very sensitive to the choice of a $T_{max}$. In particular, for large values of $T_{max}$, the contribution of the initial steps can be asymptotically washed away, resulting in a artificially low variance between different supervision. This may happen if a performance plateau is reached early. To reduce this sensitivity, we advocate for a metric averaging over multiple horizons. However, this implicitly up-weights initial steps. Formally, defining $b_i = \int_{t_i}^{t_{i+1}} m_s \, ds$ with $t_0 = 0 <$

| Control | Training set | Validation set | Metric |
|---|---|---|---|
| Depth | MultiGen-20M [4] train | MultiGen-20M [4] val | RMSE |
| Semantic Segmentation | ADE20K [20] train | ADE20K [20] val | mIoU |
| Canny Edge | ADE20K [20] train | ADE20K [20] val | F1 |
| Pose | MS-COCO [6] train filtered | MS-COCO [6] val filtered | Average Precision (AP) |
| Box+Text grounding | GoldG [3]+SBU [10]+CC3M [14]+Object365 [13] | MS-COCO [6] subset [17] | Average Precision (AP) |
| Box+Text+Image grounding | GoldG [3]+SBU [10]+CC3M [14]+Object365 [13] | MS-COCO [6] subset [17] | Average Precision (AP) |

Table 1. **Datasets used for experiments**

$t_1 < \cdots < t_n$ different training horizons, we can also write $\mathbf{mAUCC} = \sum_{i=0}^{n-1} w_i b_i$, with $w_i = \frac{1}{n} \sum_{j=i+1}^{n} \frac{1}{t_j}$, and $\mathbf{AUCC}@t_n = \sum_{i=0}^{n-1} \frac{1}{t_n} b_i$. Thus, while $\mathbf{AUCC}@T_{max}$ gives a uniform weight to all $b_i$, $\mathbf{mAUCC}$ emphasizes initial steps since $w_i$ is decreasing. This is in fact desired in our context for assessing convergence speed. For completeness we additionally report the values of $\mathbf{AUCC}@T_{max}$ on Tabs. 3 and 4 for completeness. We can see that the two metrics have similar trends.

We also provide the implementation of the mean Area Under the Convergence Curve (mAUCC) in Algorithm 1. As explained in the main paper, to obtain normalized values, we integrate the normalized convergence curves over $[0, 1]$. To do so the metric are divided by their maximum achievable value, which 255 for the RMSE and 100 for the others. We reporte the mAUCC on a $[0, 100]$ scale on the result tables.

## B. Conversion formulas

We describe the way to convert the different objectives of diffusion and flow matching to an $x_0$ estimate. The derivations are mostly reported from existing works [7, 12]. Our main contribution is to analyse their impact on controllable generation and propose a method for improving both the training speed and the final performance.

Although flow matching relies on Ordinary Differential Equations (ODEs) and diffusion modeling relies on Stochastic Differential Equations (SDEs), during training they are equivalent and the main difference is the learning objective. In particular, the two paradigms are unified under the framework of stochastic interpolants [1], they both use the one-sided interpolant:

$$x_t = \alpha_t x_0 + \sigma_t \epsilon \quad (1)$$

where $\alpha_t$ and $\sigma_t$ are respectively decreasing and increasing time-dependent functions on $[0, T]$, such that $\alpha_0 = \sigma_T = 1$ and $\alpha_T = \sigma_0 = 0$. Note that in the flow matching literature, this time convention is reversed. That is, $\alpha_t$ and $\sigma_t$ are increasing and decreasing such that $\alpha_0 = \sigma_T = 0$ and $\alpha_T = \sigma_0 = 1$. But, for notation consistency, we will use the first convention so that $x_0$ is the clean image and $x_T$ the fully noised image. The main distinction between the two paradigms comes from how $\alpha_t$ and $\sigma_t$ are instantiated and which objective is learned, each objective giving a different parameterization. Despite this variety of objectives, they can be converted into one another. As a matter of fact, the different learning losses can be seen as a particular weighting of the $x_0$ supervision loss:

$$\mathcal{L}_\theta^p = \mathbb{E}_{t, \epsilon, x_0} \left[ w_t^p \| x_0 - x_\theta(x_t, t) \|_2^2 \right] \quad (2)$$

where $w_t^p$ is the weighting for the parameterization $p$. A summary of the conversion formulas and their equivalent loss re-weightings are provided in Tab. 5. We now describe the derivations for the different parameterizations in details.

| Method | Base model | Batch size | Training steps |
|---|---|---|---|
| ControlNet [19] | Stable Diffusion 1.5 [11] | 8 | 200k |
| T2I-Adapter [9] | Stable Diffusion 1.5 [11] | 8 | 200k |
| OminiControl [16] | FLUX.1 [2] | 8 | 40k |
| GLIGEN [5] | Stable Diffusion 1.4 [11] | 64 | 200k |

Table 2. **Methods used for experiments**

**Algorithm 1** Python code for mAUCC

```python
import numpy as np
import scipy

def calculate_aucc(convergence_curve):
    # Computes the area using the trapezoidal rule
    N = len(convergence_curve)
    steps = np.arange(N) / N
    return scipy.integrate.trapezoid(
        performance_curve,
        steps
    )

def calculate_mean_aucc(
    convergence_curve,
    thresholds=None
):
    # Computes the average AUCC over multiple horizons
    if thresholds is None:
        thresholds = np.linspace(
            0.25, 1, int(np.round((1 - 0.25)/.05)) + 1,
            endpoint=True
        )
    AUCCs = []
    for ti in thresholds:
        max_steps = int(len(performance_curve) * ti)
        curve_trunc = convergence_curve[:max_steps]
        AUCC_at_ti = calculate_aucc(curve_trunc)
        AUCCs.append(AUCC_at_ti)
    mAUCC = np.mean(AUCCs)
    return mAUCC
```

| Supervision | Depth | | Semantic Seg | | Canny Edge | | Pose | |
|---|---|---|---|---|---|---|---|---|
| | **AUCC@$T_{max}$** ↓ | **mAUCC**↓ | **AUCC@$T_{max}$** ↑ | **mAUCC**↑ | **AUCC@$T_{max}$** ↑ | **mAUCC**↑ | **AUCC@$T_{max}$** ↑ | **mAUCC**↑ |
| ControlNet (Diffusion) | | | | | | | | |
| $\epsilon$-ControlNet | 15.97 | 17.70 | 30.61 | 25.19 | 9.24 | 8.09 | 44.85 | 35.86 |
| $v$-ControlNet | 18.56 | 21.89 | 20.39 | 13.48 | 8.41 | 6.93 | 36.20 | 22.72 |
| $x_0$-ControlNet | **15.01** | **15.98** | **35.60** | **31.52** | **9.71** | **8.76** | **48.63** | **42.19** |
| T2I-Adapter (Diffusion) | | | | | | | | |
| $\epsilon$-T2I-Adpater | 21.77 | 22.44 | 19.00 | 17.39 | 3.73 | 3.61 | 15.66 | 12.98 |
| $v$-T2I-Adapter | 25.94 | 26.52 | 12.51 | 11.54 | 2.66 | 2.62 | 6.02 | 4.44 |
| $x_0$-T2I-Adapter | **19.53** | **19.67** | **22.01** | **21.07** | **4.51** | **4.47** | **23.00** | **21.45** |
| OminiControl (Flow Matching) | | | | | | | | |
| $u$-OminiControl | **13.38** | **13.47** | 41.37 | **39.13** | **15.05** | 14.40 | 43.87 | 34.54 |
| $\epsilon$-OminiControl | 15.45 | 16.79 | 29.28 | 23.76 | 11.15 | 9.71 | 26.78 | 17.57 |
| $x_0$-OminiControl | 13.58 | 13.74 | **41.39** | 38.69 | 14.99 | **14.41** | **49.28** | **42.07** |

Table 3. **Comparison between the different supervision signals on different spatially-aligned control modalities and methods.** The curves are smoothed with EMA weight 0.9 before computing **AUCC@$T_{max}$** and **mAUCC**.

| Supervision | Box+Text | | Box+Text+Image | |
|---|---|---|---|---|
| | **AUCC@$T_{max}$** ↑ | **mAUCC**↑ | **AUCC@$T_{max}$** ↑ | **mAUCC**↑ |
| GLIGEN (Diffusion) | | | | |
| $\epsilon$-GLIGEN | 14.61 | 8.28 | 9.92 | 6.15 |
| $v$-GLIGEN | 2.07 | 1.54 | 3.34 | 9.56 |
| $x_0$-GLIGEN | **18.38** | **18.38** | **11.49** | 8.07 |

Table 4. **Comparison between the different supervision signals on different non-spatially-aligned control modalities.** The curves are smoothed with EMA weight 0.9 before computing **AUCC@$T_{max}$** and **mAUCC**.

| Parameterization | Conversion | Loss re-weighting |
|---|---|---|
| $\epsilon$ | $\frac{1}{\alpha_t}, -\frac{\sigma_t}{\alpha_t}$ | $\frac{\sigma_t^2}{\alpha_t^2}$ |
| $v$ | $\alpha_t, -\sigma_t$ | $\left(\frac{\alpha_t^2}{\sigma_t^2}+1\right)^{-1}$ |
| $u$ | $-\frac{\dot{\sigma}_t}{\dot{\alpha}_t\sigma_t - \alpha_t\dot{\sigma}_t}, \frac{\sigma_t}{\dot{\alpha}_t\sigma_t - \alpha_t\dot{\sigma}_t}$ | $\left(\frac{\sigma_t}{\dot{\alpha}_t\sigma_t - \alpha_t\dot{\sigma}_t}\right)^2$ |

Table 5. **Conversion formulas from different predictors to an $x_0$-predictor**. In the second column, $(a_t, b_t)$ corresponds to $x_\theta(x_t, t) = a_t x_t + b_t f(x_t, t)$ where $f$ is the predictor. In the third column are the equivalent re-weighting factors of the initial losses.

## B.1. Diffusion

In diffusion, $\alpha_t$ and $\sigma_t$ are indirectly defined by reasoning on the forward diffusion process.

$\epsilon$-**parameterization.** Using the relation Eq. (1) one can write:

$$x_0 = \frac{x_t - \sigma_t \epsilon}{\alpha_t} \text{ and } \epsilon = \frac{x_t - \alpha_t x_0}{\sigma_t} \quad (3)$$

hence the loss of an $\epsilon$-predictor can be written as:

$$\mathcal{L}_\theta^\epsilon = \mathbb{E}_{t,\epsilon,x_0}\left[\|\epsilon - \epsilon_\theta(x_t, t)\|_2^2\right], \quad (4)$$

$$= \mathbb{E}_{t,\epsilon,x_0}\left[\|\frac{1}{\sigma_t}(x_t - \alpha_t x_0) - \frac{1}{\sigma_t}(x_t - \alpha_t x_\theta(x_t, t))\|_2^2\right] \quad (5)$$

$$= \mathbb{E}_{t,\epsilon,x_0}\left[\frac{\alpha_t^2}{\sigma_t^2}\|x_0 - x_\theta(x_t, t)\|_2^2\right] \quad (6)$$

hence $w_t^\epsilon = \frac{\alpha_t^2}{\sigma_t^2}$, which is the signal-to-noise ratio (SNR).

$v$-**parameterization.** [12] defines the $v$-parameterization, also known as the Variance Preserving (VP) schedule, in the case where $\alpha_t^2 + \sigma_t^2 = 1$. Under this assumption, the interpolant is expressed in terms of $\phi_t = \arctan(\sigma_t/\alpha_t)$:

$$x_\phi = \alpha_\phi x_0 + \sigma_\phi \epsilon = \cos(\phi) x_0 + \sin(\phi) \epsilon \quad (7)$$

hence $v$ is defined as:

$$v_\phi = \frac{dx_\phi}{d\phi} = \cos(\phi)\epsilon - \sin(\phi)x_0 \quad (8)$$

Therefore, $v$ is an angular velocity. We can then derive $x_0$ from $v$ as follows:

$$x_0 = \frac{\cos(\phi)\epsilon - v_\phi}{\sin(\phi)} \quad (9)$$

$$= \frac{\cos(\phi)\frac{x_\phi - \cos(\phi)x_0}{\sin(\phi)} - v_\phi}{\sin(\phi)} \quad (10)$$

$$= \frac{\cos(\phi)x_\phi - \cos^2(\phi)x_0 - \sin(\phi)v_\phi}{\sin^2(\phi)} \quad (11)$$

$$\sin^2(\phi)x_0 = \cos(\phi)x_\phi - \cos^2(\phi)x_0 - \sin(\phi)v_\phi \quad (12)$$

$$\left[\cos^2(\phi) + \sin^2(\phi)\right]x_0 = \cos(\phi)x_\phi - \sin(\phi)v_\phi \quad (13)$$

$$x_0 = \cos(\phi)x_\phi - \sin(\phi)v_\phi \quad (14)$$

hence, coming back to $\alpha_t$ and $\sigma_t$, we can summarize by writing:

$$x_0 = \alpha_t x_t - \sigma_t v_t \text{ and } v_t = \alpha_t \epsilon - \sigma_t x_0 \quad (15)$$

Using these expressions, the loss of a $v$-predictor can be written as:

$$\mathcal{L}_\theta^v = \mathbb{E}_{t,\epsilon,x_0}\left[\|v_t - v_\theta(x_t,t)\|_2^2\right], \quad (16)$$

$$= \mathbb{E}_{t,\epsilon,x_0}\left[\|\frac{1}{\sigma_t}(\alpha_t x_t - x_0) - \frac{1}{\sigma_t}(\alpha_t x_t - x_\theta(x_t,t))\|_2^2\right] \quad (17)$$

$$= \mathbb{E}_{t,\epsilon,x_0}\left[\frac{1}{\sigma_t^2}\|x_0 - x_\theta(x_t,t)\|_2^2\right] \quad (18)$$

hence, $w_t^v = \frac{1}{\sigma_t^2} = \frac{\alpha_t^2 + \sigma_t^2}{\sigma_t^2} = \frac{\alpha_t^2}{\sigma_t^2} + 1$, which is referred to as SNR+1 weighting in [12], where the authors used it in the context of diffusion model distillation.

### B.2. Flow matching

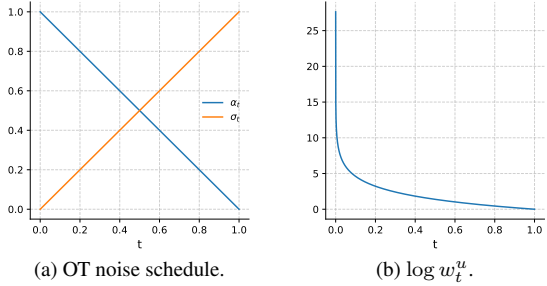

Figure 1. **OT noise schedule used in flow matching and the corresponding weighting incurred by $x_0$-supervision in log scale.**

In flow matching, the goal is to predict the velocity field $u$ governing the probability flow ODE [15]:

$$\frac{dX_t}{dt} = u_t(X_t) \quad (19)$$

Equation (1) describes the conditional flow $\psi_t(x_0, \epsilon) : t \mapsto \alpha_t x_0 + \sigma_t \epsilon$, and one has:

$$u_t(x) = \mathbb{E}_{x_0, \epsilon}[u_t(X_t|x_0, \epsilon)|X_t = x] \quad (20)$$

$$= \mathbb{E}_{x_0, \epsilon}\left[\dot{\psi}_t(x_0, \epsilon)\Big|X_t = x\right] \quad (21)$$

$$= \mathbb{E}_{x_0, \epsilon}[\dot{\alpha}_t x_0 + \dot{\sigma}_t \epsilon | X_t = x] \quad (22)$$

$$= \mathbb{E}_{x_0, \epsilon}\left[\dot{\alpha}_t x_0 + \dot{\sigma}_t \frac{X_t - \alpha_t x_0}{\sigma_t}\Big| X_t = x\right] \quad (23)$$

$$= \mathbb{E}_{x_0, \epsilon}\left[\frac{\dot{\alpha}_t \sigma_t - \alpha_t \dot{\sigma}_t}{\sigma_t} x_0 + \frac{\dot{\sigma}_t}{\sigma_t} x \Big| X_t = x\right] \quad (24)$$

$$= \frac{\dot{\alpha}_t \sigma_t - \alpha_t \dot{\sigma}_t}{\sigma_t}\underbrace{\mathbb{E}_{x_0, \epsilon}[x_0|X_t = x]}_{x_0\text{-predictor}} + \frac{\dot{\sigma}_t}{\sigma_t} x \quad (25)$$

where the $\cdot$ denotes the time derivative and Eq. (23) uses the fact that $\epsilon = \frac{X_t - \alpha_t x_0}{\sigma_t}$. In practice, one trains a network to predict the conditional velocity field $u_t(\cdot|x_0, \epsilon)$ by optimizing the conditional flow matching loss:

$$\mathcal{L}_\theta^u = \mathbb{E}_{t,x_0,\epsilon}\left[\|u_t(x_t|x_0,\epsilon) - u_\theta(x_t,t)\|_2^2\right] \quad (26)$$

$$= \mathbb{E}_{t,x_0,\epsilon}\left[\left(\frac{\dot{\alpha}_t \sigma_t - \alpha_t \dot{\sigma}_t}{\sigma_t}\right)^2 \|x_0 - x_\theta(x_t,t)\|_2^2\right] \quad (27)$$

where Eq. (27) uses the relation between the velocity and $x_0$ in Eqs. (24) and (25). Hence, we get $w_t^u = \left(\frac{\dot{\alpha}_t \sigma_t - \alpha_t \dot{\sigma}_t}{\sigma_t}\right)^2$. This can be re-written as follows:

$$w_t^u = \left(\alpha_t\left(\frac{\dot{\alpha}_t}{\alpha_t} - \frac{\dot{\sigma}_t}{\sigma_t}\right)\right)^2 \quad (28)$$

$$= \alpha_t^2\left(\frac{d\log\alpha}{dt}(t) - \frac{d\log\sigma}{dt}(t)\right)^2 \quad (29)$$

$$= \alpha_t^2\left(\frac{d}{dt}\left(\log\frac{\alpha}{\sigma}\right)(t)\right)^2 \quad (30)$$

$$= \frac{\alpha_t^2}{4}\left(\frac{d\log\text{SNR}}{dt}(t)\right)^2 \quad (31)$$

Although theoretically $\left(\frac{d\log\text{SNR}(t)}{dt}\right)^2$ diverges as $t \to T$, this issue can be mitigated in practice. Overall, $w_t^u$ is very small near $T$ compared to its values near 0. Figure 1 illustrates this for the Optimal Transport (OT) path, i.e., where $\alpha_t = 1 - t$ and $\sigma_t = t$. The over-weighting of the high-SNR region relative to the low-SNR region causes the network to focus less on the early denoising steps during training.

### C. Convergence curves for T2I-Adapter

The convergence curves for T2I-Adapter are provided on Fig. 2. We can see that $x_0$-supervision consistently improves convergence on all selected tasks.

### D. Analysis of the results for converting to $v/\epsilon$

In our experiments, we observed that $v$-ControlNet and $\epsilon$-OminiControl do not achieve optimal performance. We show in this section that this can be explained by the implied weightings. Formally, in the case of $\epsilon$-to-$v$, we obtain:

$$\mathcal{L}_\theta^{\epsilon \to v} = \mathbb{E}_{t,\epsilon,x_0}\left[\|v_t - (\alpha_t \epsilon_\theta(x_t,t) - \sigma_t x_0)\|_2^2\right] \quad (32)$$

$$= \mathbb{E}_{t,\epsilon,x_0}\left[\|(\alpha_t \epsilon - \sigma_t x_0) - (\alpha_t \epsilon_\theta(x_t,t) - \sigma_t x_0)\|_2^2\right] \quad (33)$$

$$= \mathbb{E}_{t,\epsilon,x_0}\left[\alpha_t^2\|\epsilon - \epsilon_\theta(x_t,t)\|_2^2\right] \quad (34)$$

$$= \mathbb{E}_{t,\epsilon,x_0}\left[\alpha_t^2 w_t^\epsilon\|x_0 - x_\theta(x_t,t)\|_2^2\right] \quad (35)$$

$$= \mathbb{E}_{t,\epsilon,x_0}\left[\frac{\alpha_t^4}{\sigma_t^2}\|x_0 - x_\theta(x_t,t)\|_2^2\right] \quad (36)$$

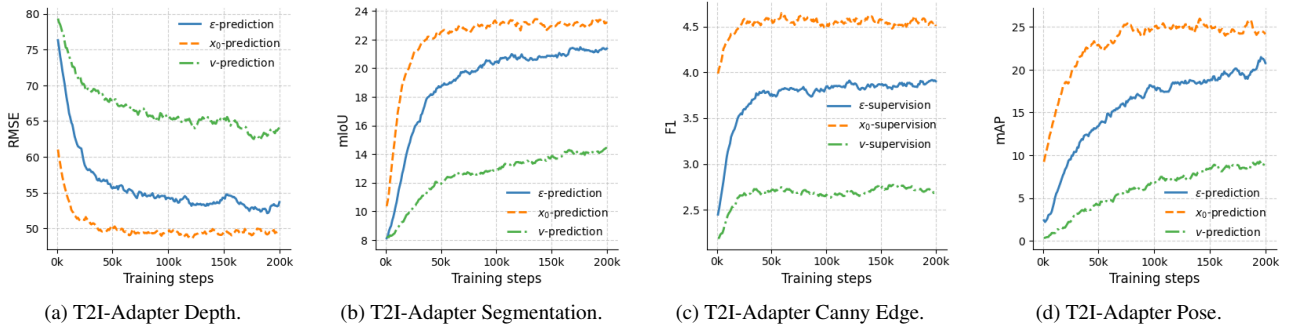

(a) T2I-Adapter Depth.   (b) T2I-Adapter Segmentation.   (c) T2I-Adapter Canny Edge.   (d) T2I-Adapter Pose.

Figure 2. **Convergence curves for T2I-Adapter on different tasks.** We use an EMA weight of 0.9 to smooth the curves. We can notice that the convergence is faster with $x_0$-supervision.

and for the case of $u$-to-$\epsilon$ we get:

$$\mathcal{L}_\theta^{u \to \epsilon} = \mathbb{E}_{t,\epsilon,x_0}\left[\|\epsilon - \frac{\alpha_t}{\alpha_t \dot{\sigma}_t - \dot{\alpha}_t \sigma_t}(u_\theta(x_t,t) - \frac{\dot{\alpha}_t}{\alpha_t}x_t)\|_2^2\right] \quad (37)$$

$$= \mathbb{E}_{t,\epsilon,x_0}\left[\left(\frac{\alpha_t}{\alpha_t \dot{\sigma}_t - \dot{\alpha}_t \sigma_t}\right)^2 \|u_t(x_t|x_0,\epsilon) - u_\theta(x_t,t)\|_2^2\right] \quad (38)$$

$$= \mathbb{E}_{t,\epsilon,x_0}\left[\left(\frac{\alpha_t}{\alpha_t \dot{\sigma}_t - \dot{\alpha}_t \sigma_t}\right)^2 w_t^u \|x_0 - x_\theta(x_t,t)\|_2^2\right] \quad (39)$$

$$= \mathbb{E}_{t,\epsilon,x_0}\left[\frac{\alpha_t^2}{\sigma_t^2}\|x_0 - x_\theta(x_t,t)\|_2^2\right] \quad (40)$$

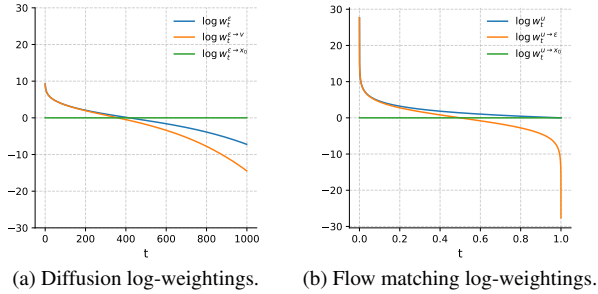

(a) Diffusion log-weightings.   (b) Flow matching log-weightings.

Figure 3. **Log-weighting functions.** In Fig. 3a, we plot the log-weightings incurred by the $x_0$-supervision loss when using $\epsilon$, $v$, and $x_0$ as supervision signals for SD-based control methods. In Fig. 3b, we plot the corresponding weightings when using $u$, $\epsilon$, and $x_0$ as supervision signals for OminiControl. In both cases, we observe that lower convergence speed and performance are related to suboptimal weighting of the initial denoising steps (near $T$). In the case of $u$-supervision in particular, the weighting converges to 1 near $T$, making it less detrimental than $\epsilon$-supervision. However, since its weighting near 0 is significantly higher, the training gradients focus more on these steps, which are less critical for the final image's control fidelity than the initial denoising steps.

hence both result in worse weightings that penalizes low SNRs. Therefore resulting in a slower convergence and worse performances in chosen training horizons. This is illustrated on Fig. 3.

## E. Qualitative results

On Figs. 4 to 17, we provide some visual results on the control fidelity. Overall, with $x_0$-supervision, we either maintain or significantly improve the control fidelity.

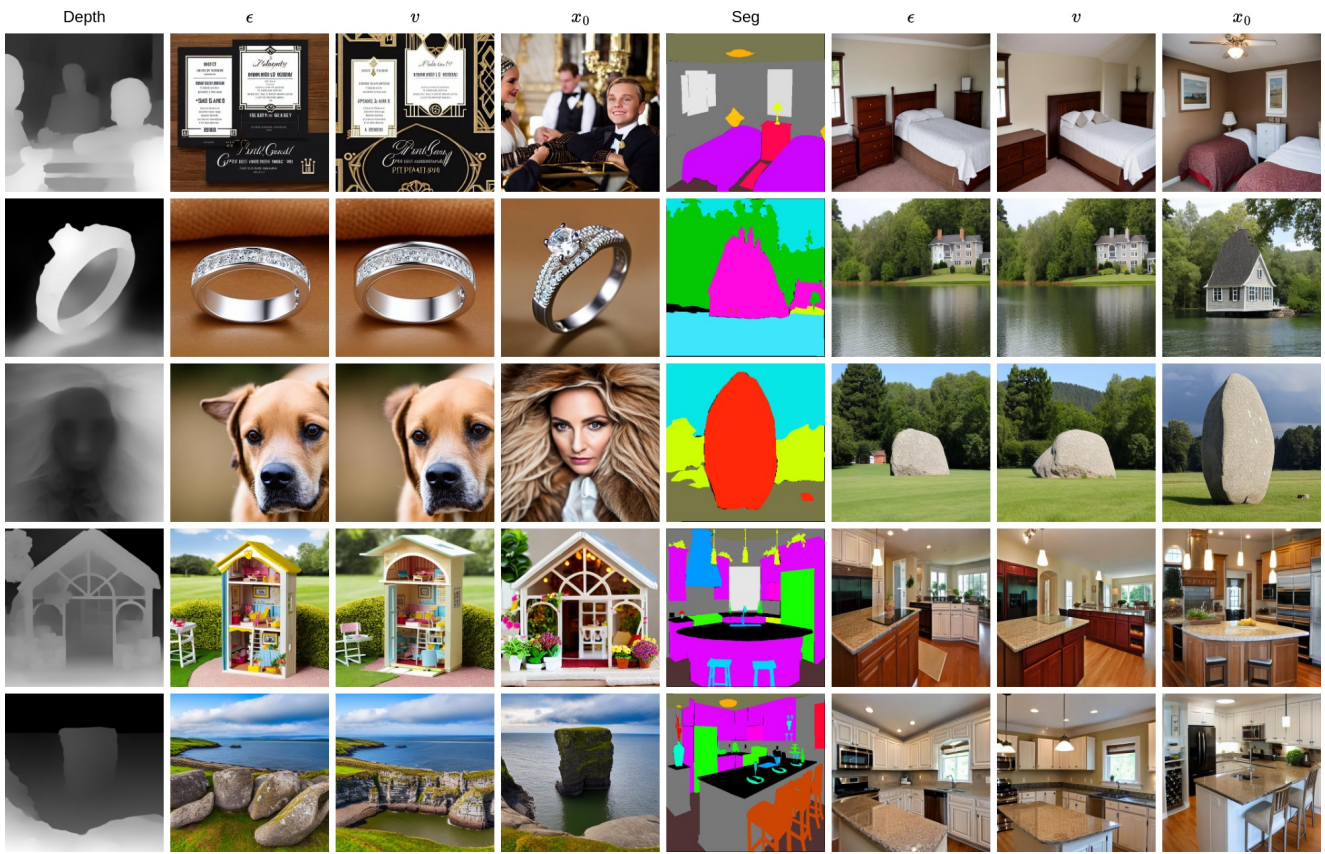

Figure 4. Qualitative results on depth and segmentation ControlNet with the three supervision signals after 10k training steps.

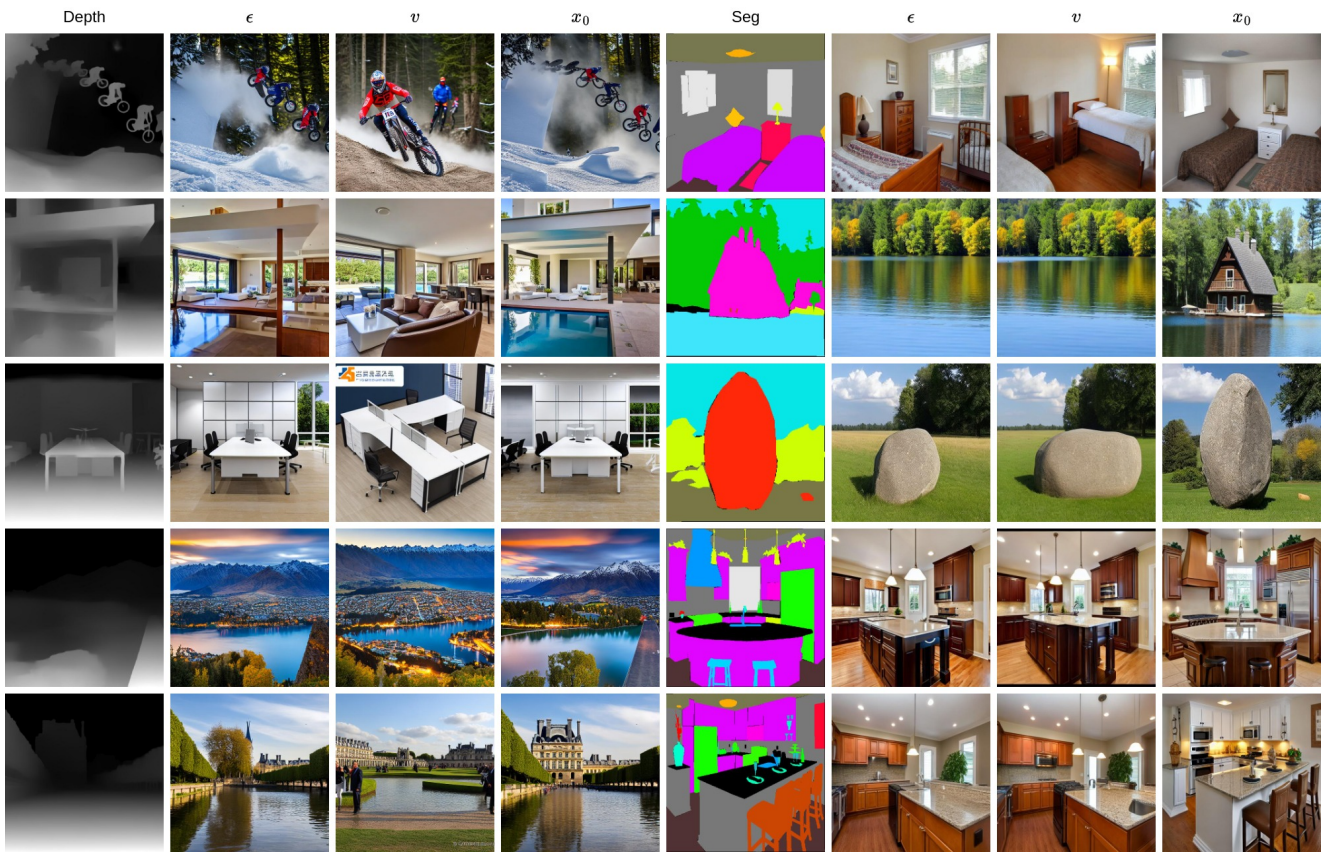

Figure 5. Qualitative results on depth and segmentation ControlNet with the three supervision signals after 25k training steps.

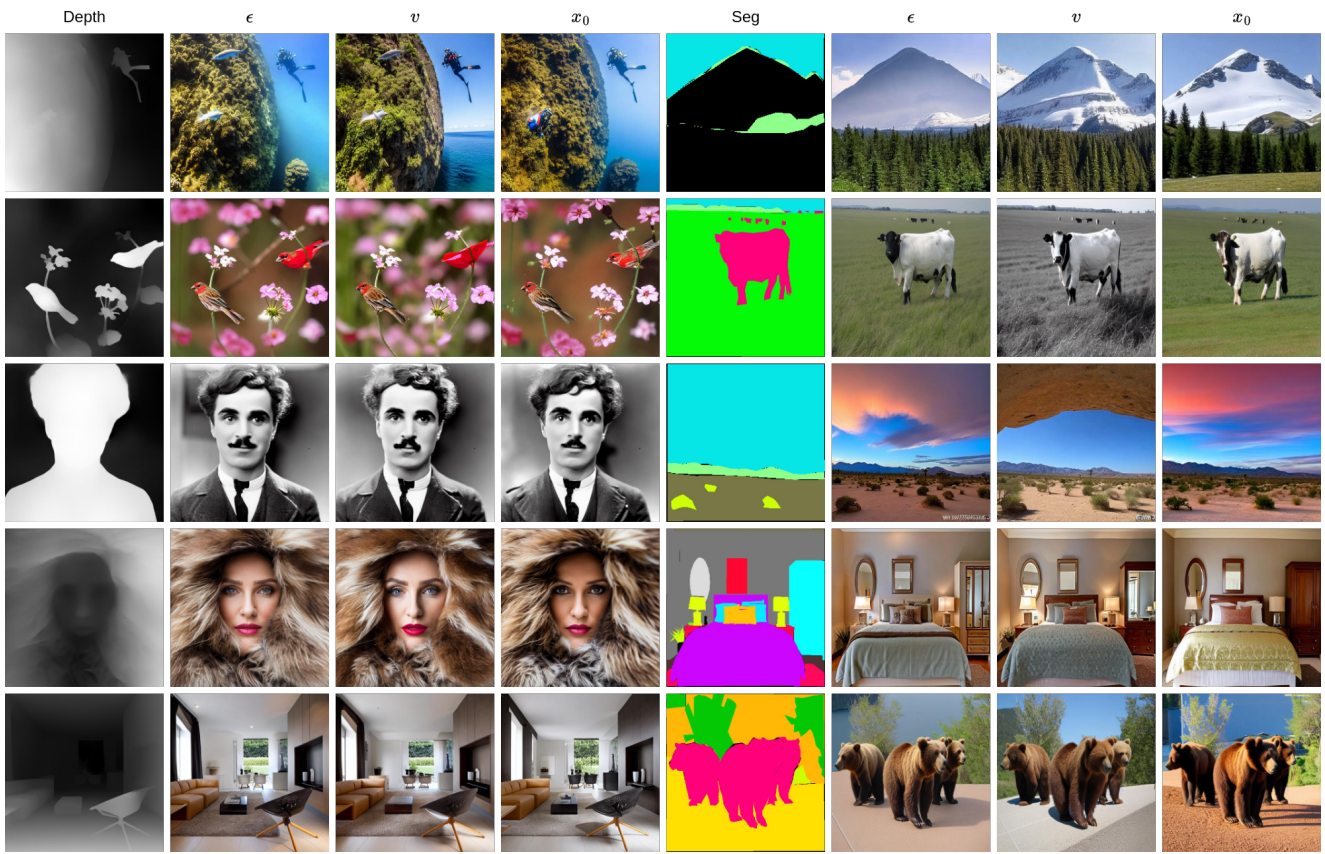

Figure 6. Qualitative results on depth and segmentation ControlNet with the three supervision signals after 200k training steps.

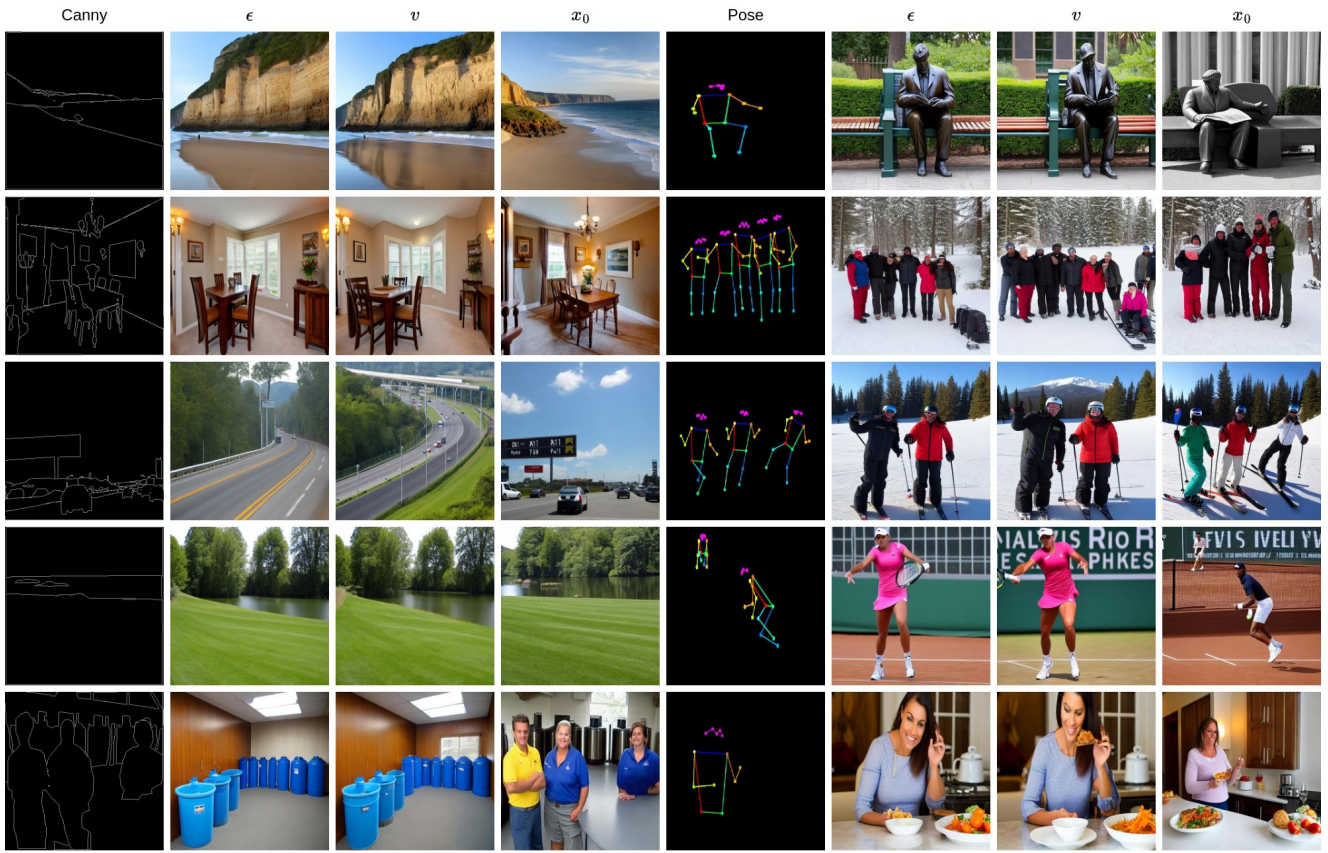

Figure 7. Qualitative results on Canny and pose ControlNet with the three supervision signals after 10k training steps.

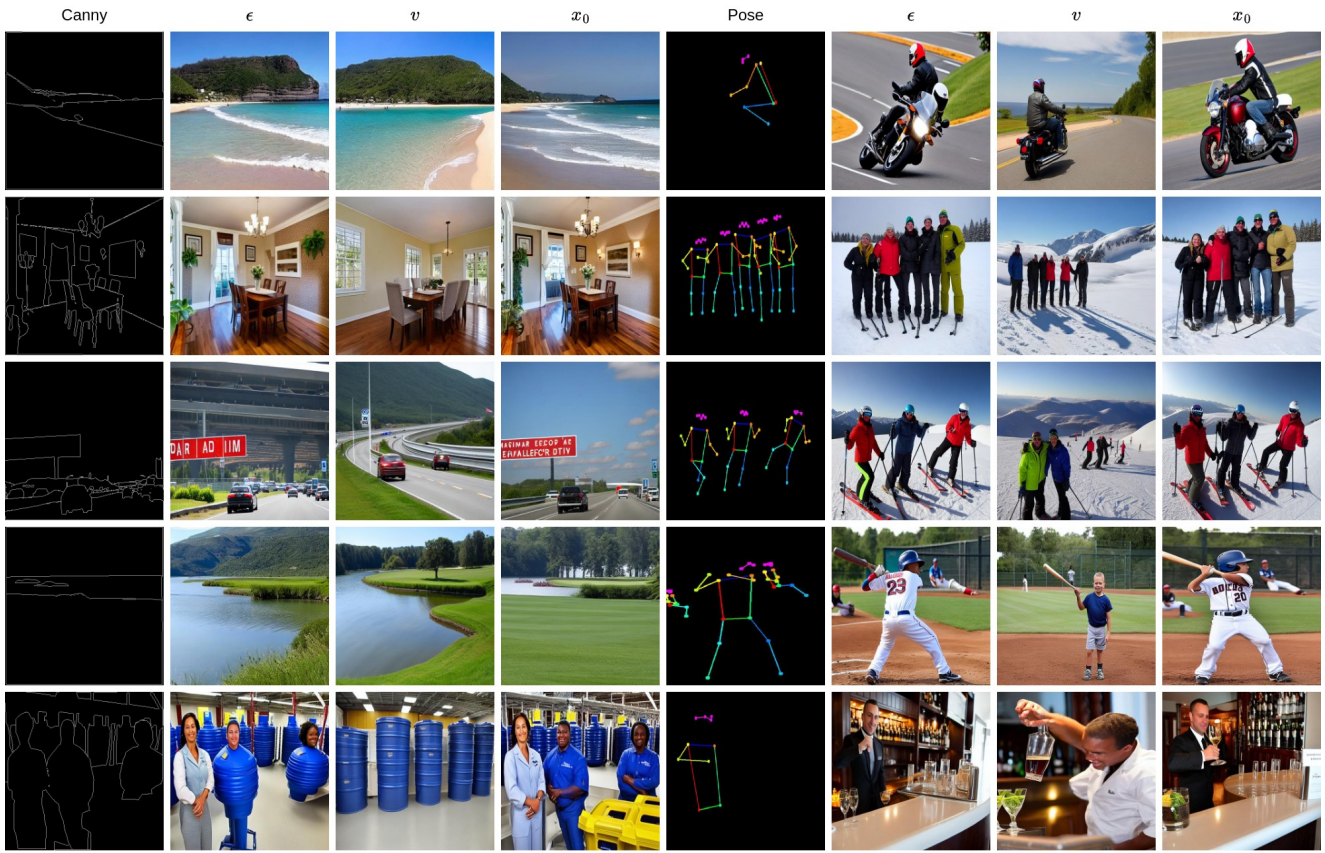

Figure 8. Qualitative results on Canny and pose ControlNet with the three supervision signals after 25k training steps.

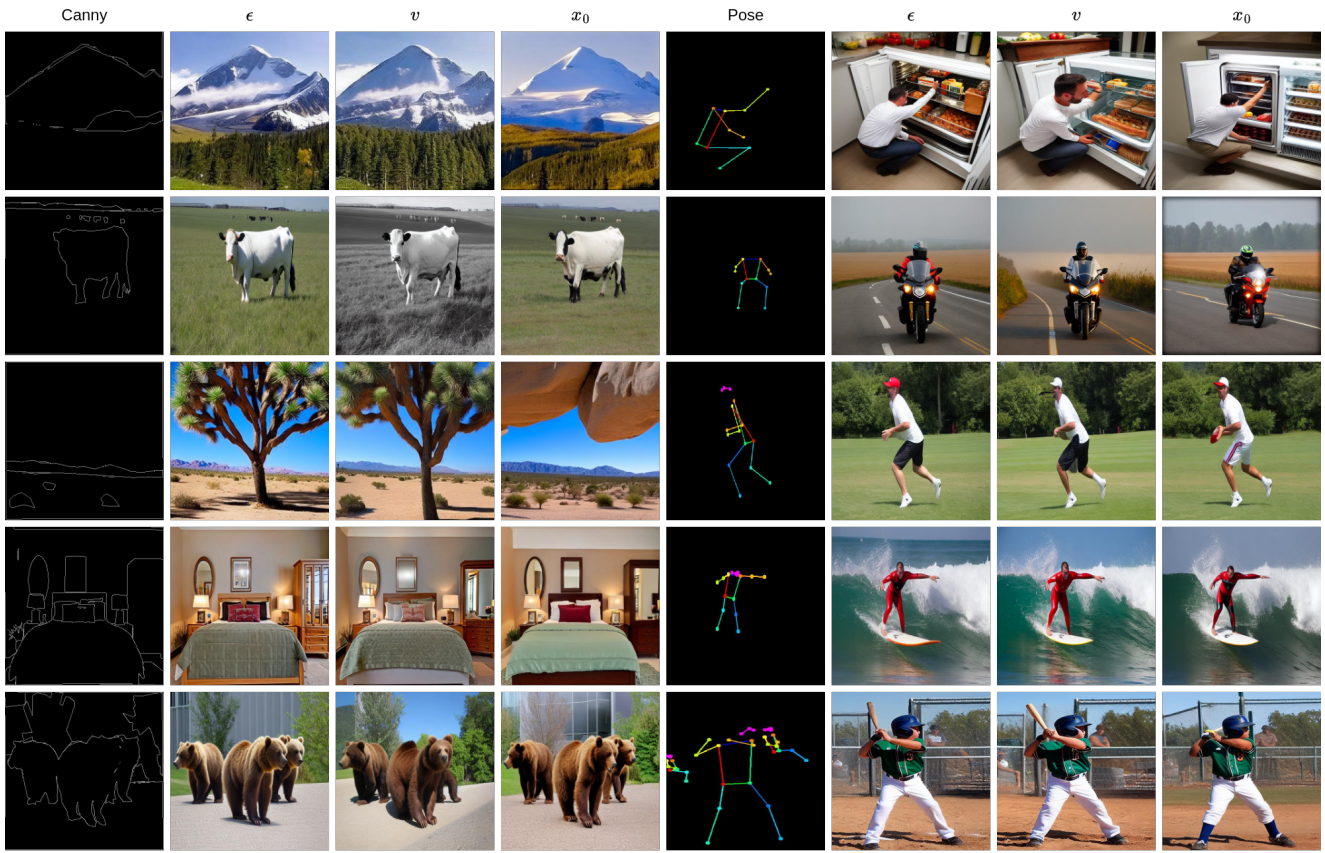

Figure 9. Qualitative results on Canny edge and pose ControlNet with the three supervision signals after 200k training steps.

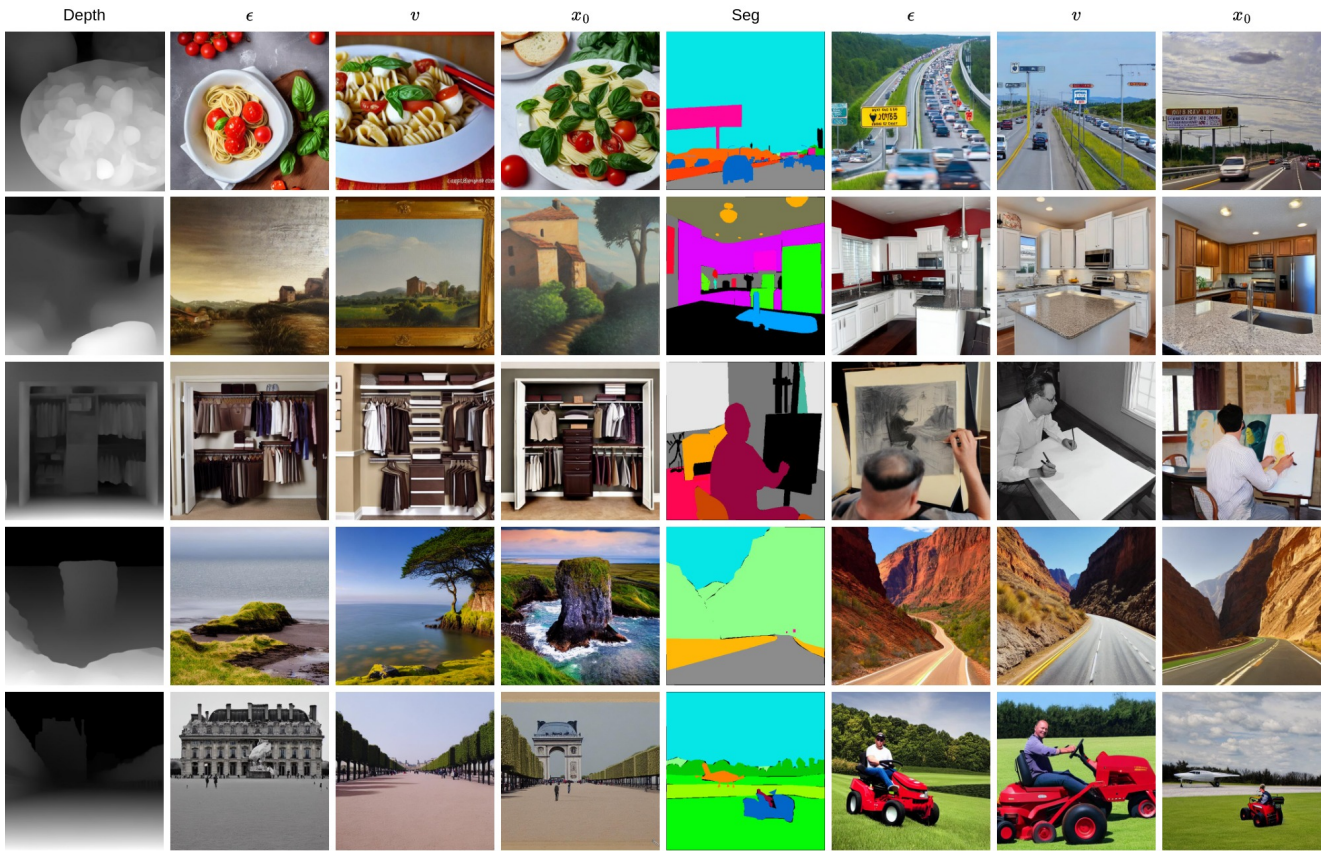

Figure 10. Qualitative results on depth and segmentation T2I-Adapter with the three supervision signals after 10k training steps.

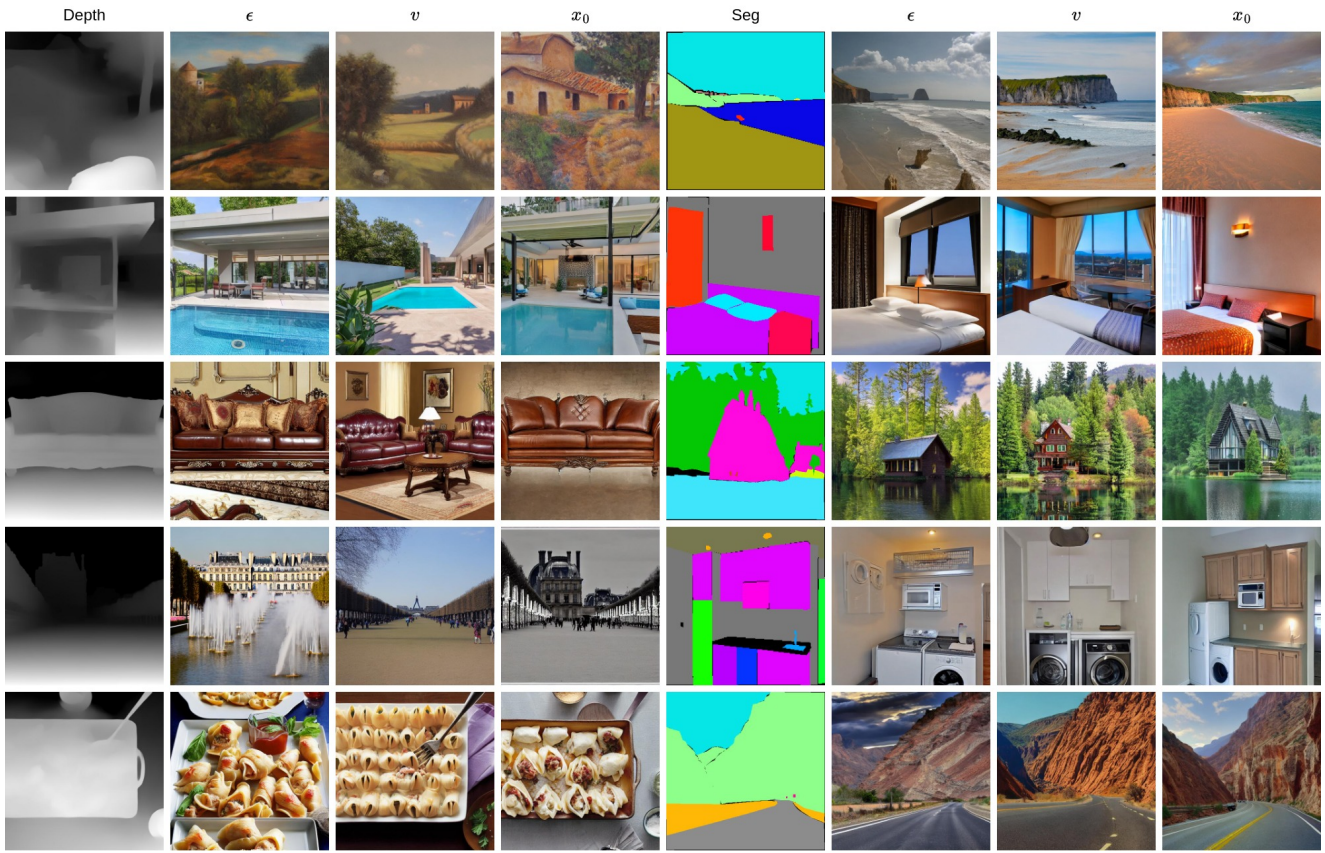

Figure 11. Qualitative results on depth and segmentation T2I-Adapter with the three supervision signals after 200k training steps.

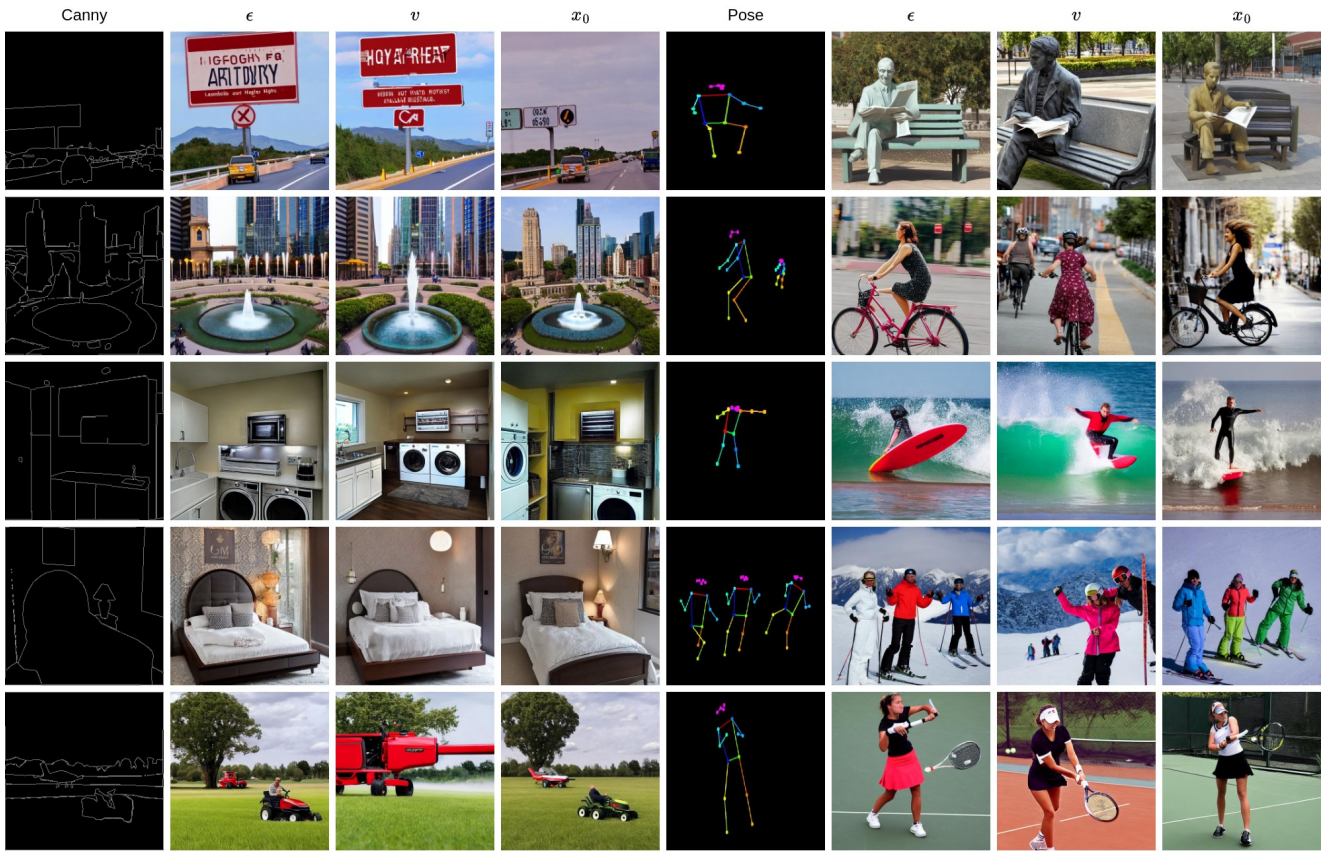

Figure 12. Qualitative results on Canny edge and pose T2I-Adapter with the three supervision signals after 10k training steps.

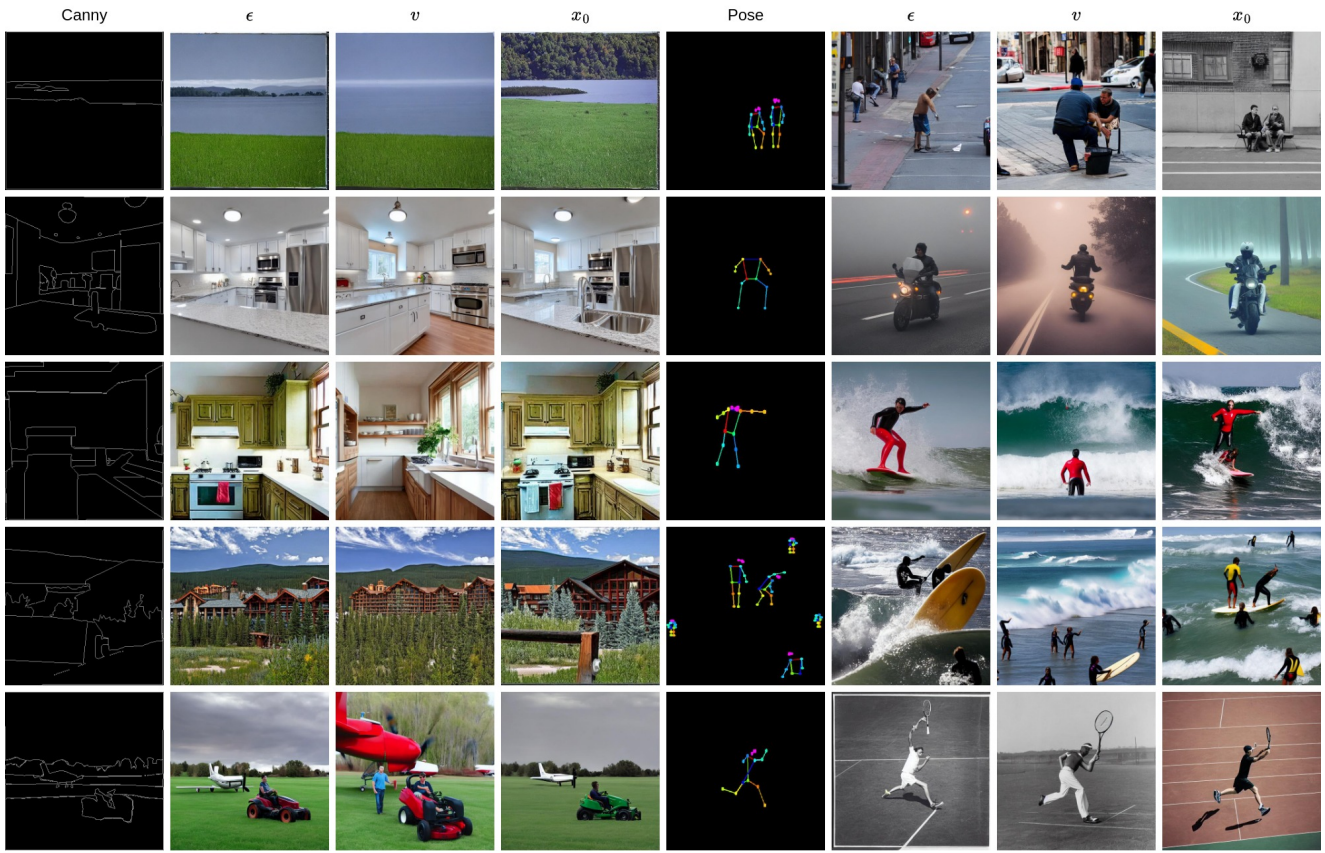

Figure 13. Qualitative results on Canny edge and pose T2I-Adapter with the three supervision signals after 200k training steps.

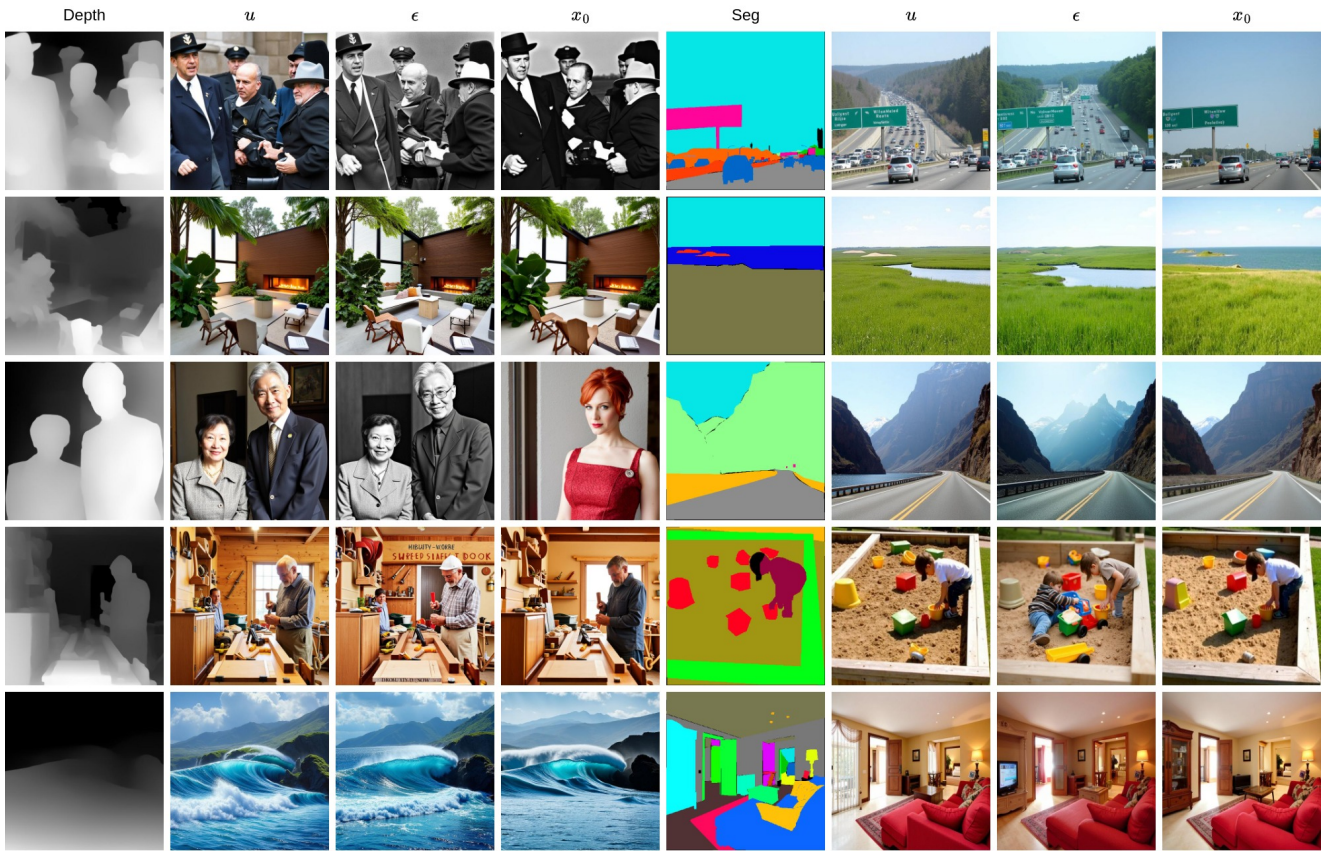

Figure 14. Qualitative results on depth and segmentation OminiControl with the three supervision signals after 5k training steps.

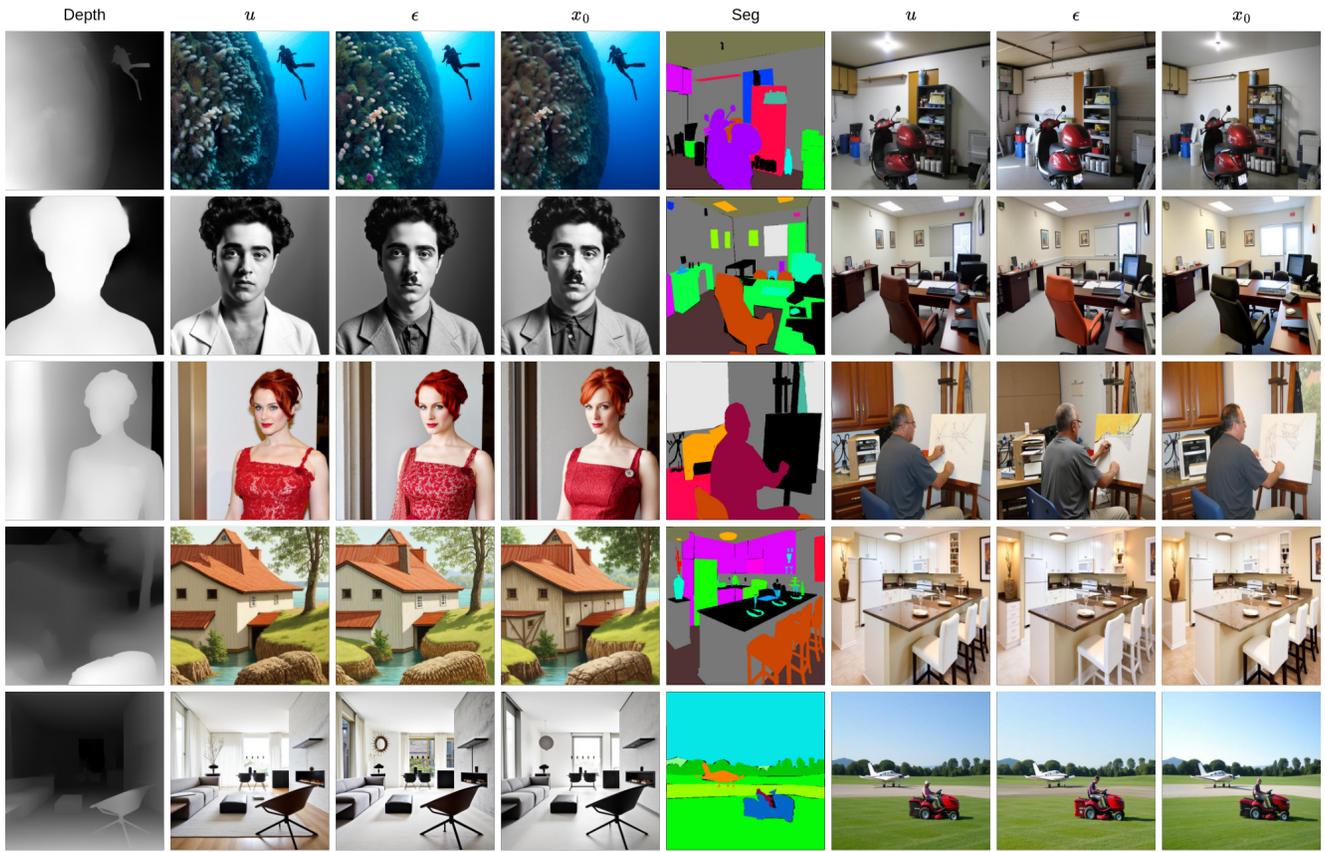

Figure 15. Qualitative results on depth and segmentation OminiControl with the three supervision signals after 40k training steps.

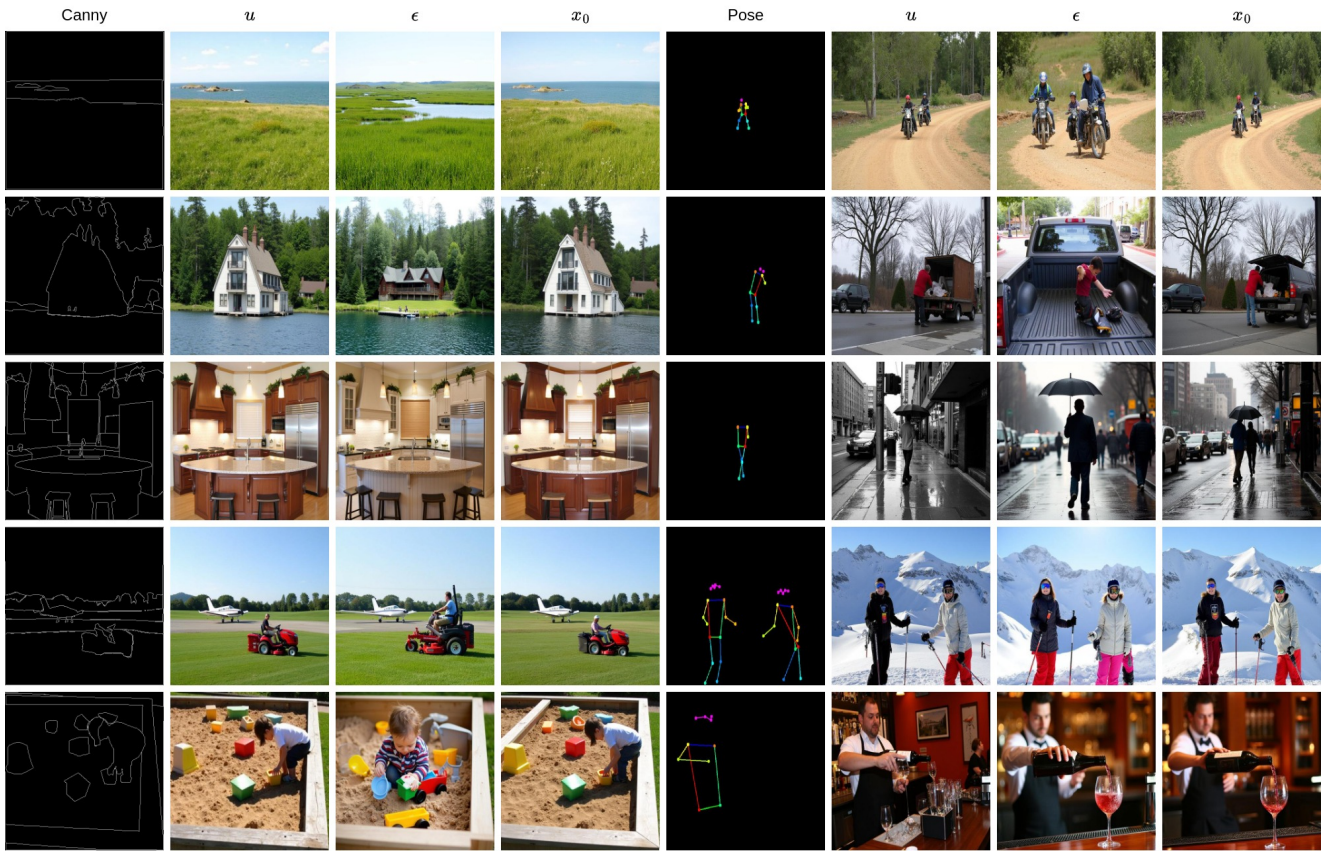

Figure 16. Qualitative results on Canny edge and pose OminiControl with the three supervision signals after 5k training steps.

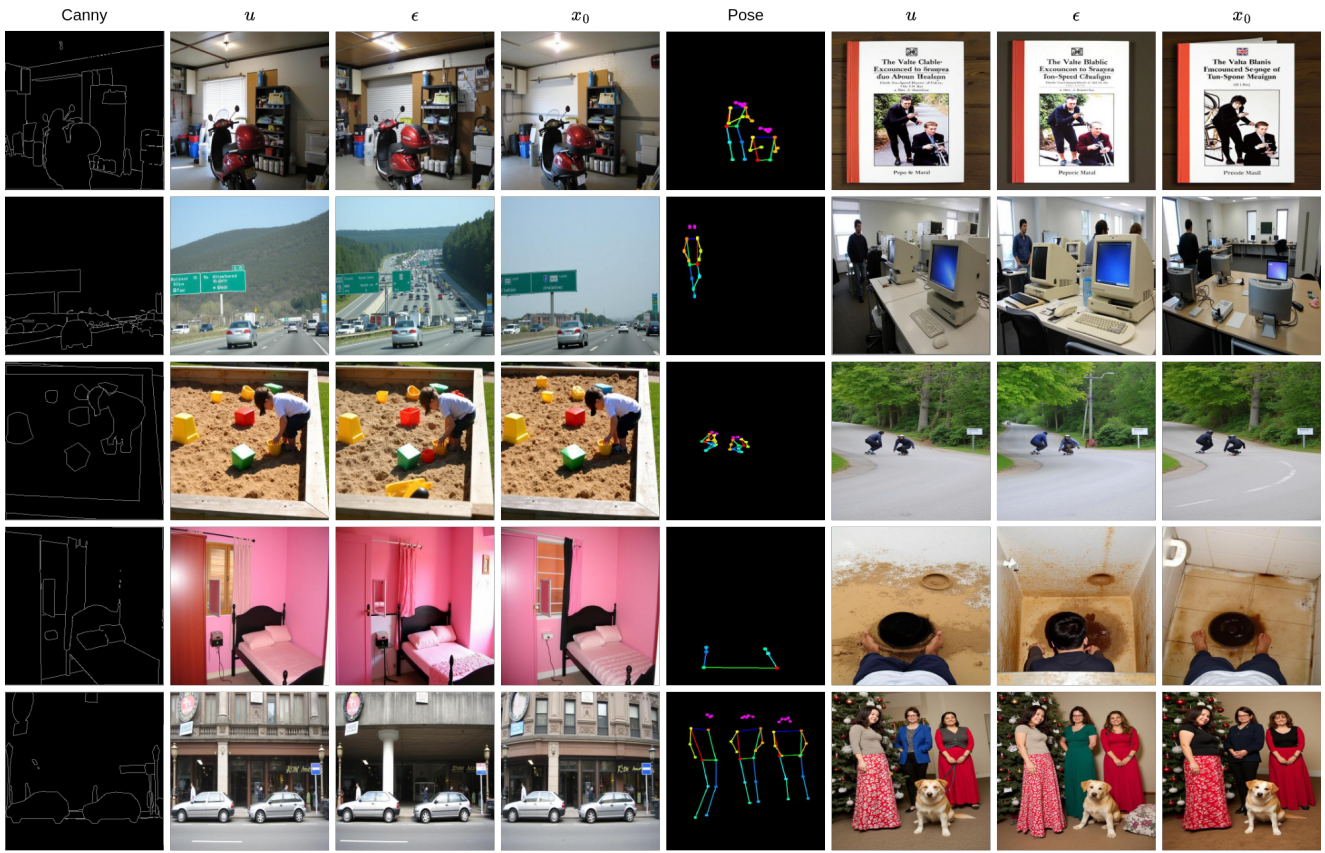

Figure 17. Qualitative results on Canny edge and pose OminiControl with the three supervision signals after 40k training steps.




\end{document}